\documentclass[lettersize,journal]{IEEEtran}
\usepackage{amsmath,amsfonts}
\usepackage{algorithmic}
\usepackage{algorithm}
\usepackage{array}
\usepackage[caption=false,font=normalsize,labelfont=sf,textfont=sf]{subfig}
\usepackage{textcomp}
\usepackage{stfloats}
\usepackage{url}
\usepackage{verbatim}
\usepackage{graphicx}
\usepackage{cite}
\usepackage{multirow}
\usepackage{xspace}
\usepackage{textcomp}
\usepackage{gensymb}
\usepackage{amssymb}
\usepackage{algorithmic}
\usepackage{algorithm}
\usepackage[colorlinks,linkcolor=red,anchorcolor=blue,citecolor=green]{hyperref}
\usepackage[table, dvipsnames]{xcolor}

\newcommand*{\ie}{i.e.\@\xspace}
\newcommand{\etal}{{et al}.\@ }

\makeatletter
\newcommand*{\etc}{%
    \@ifnextchar{.}%
        {etc}%
        {etc.\@\xspace}%
}

\begin{document}

\title{Anti-Forgetting Adaptation for Unsupervised Person Re-identification}

\author{Hao Chen,
        Francois Bremond,
        Nicu Sebe,
        Shiliang Zhang
\thanks{
This work was supported in part by the Natural Science Foundation of China under Grant No. 62402013, U20B2052, 61936011, in part by Grant No. 2023-JCJQ-LA-001-088, in part by the China Postdoctoral Science Foundation under Grant No. 2023M730056, in part by the Okawa Foundation Research Award, in part by the Ant Group Research Fund, in part by the Kunpeng\&Ascend Center of Excellence, Peking University, in part by the MUR PNRR project FAIR (PE00000013) funded by the NextGenerationEU, in part by the EU Horizon project ELIAS (No. 101120237), and in part by the French government, through the 3IA Côte d’Azur Investments in the Future project managed by the National Research Agency (ANR) with the reference number ANR-19-P3IA-0002.

\IEEEcompsocthanksitem H.~Chen and S.~Zhang are with Peking University, No.5 Yiheyuan Road, Beijing 100871, China. E-mail: \{hchen, slzhang.jdl\}@pku.edu.cn
\IEEEcompsocthanksitem F.~Bremond is with Inria, 2004 Route des Lucioles, 06902 Valbonne, France. E-mail: francois.bremond@inria.fr
\IEEEcompsocthanksitem N.~Sebe is with University of Trento, Via Sommarive 9 - 38123 Povo - Trento, Italy.E-mail: niculae.sebe@unitn.it
}


\thanks{The source code will be made available at \href{https://github.com/chenhao2345/DJAA}{https://github.com/chenhao2345/DJAA}.}
}




\maketitle

\begin{abstract}

Regular unsupervised domain adaptive person re-identification (ReID) focuses on adapting a model from a source domain to a fixed target domain. However, an adapted ReID model can hardly retain previously-acquired knowledge and generalize to unseen data. In this paper, we propose a Dual-level Joint Adaptation and Anti-forgetting (DJAA) framework, which incrementally adapts a model to new domains without forgetting source domain and each adapted target domain. We explore the possibility of using prototype and instance-level consistency to mitigate the forgetting during the adaptation. Specifically, we store a small number of representative image samples and corresponding cluster prototypes in a memory buffer, which is updated at each adaptation step. With the buffered images and prototypes, we regularize the image-to-image similarity and image-to-prototype similarity to rehearse old knowledge. After the multi-step adaptation, the model is tested on all seen domains and several unseen domains to validate the generalization ability of our method. Extensive experiments demonstrate that our proposed method significantly improves the anti-forgetting, generalization and backward-compatible ability of an unsupervised person ReID model. 
\end{abstract}


\section{Introduction}
\begin{figure}[t]
\centering
   \includegraphics[width=0.95\linewidth]{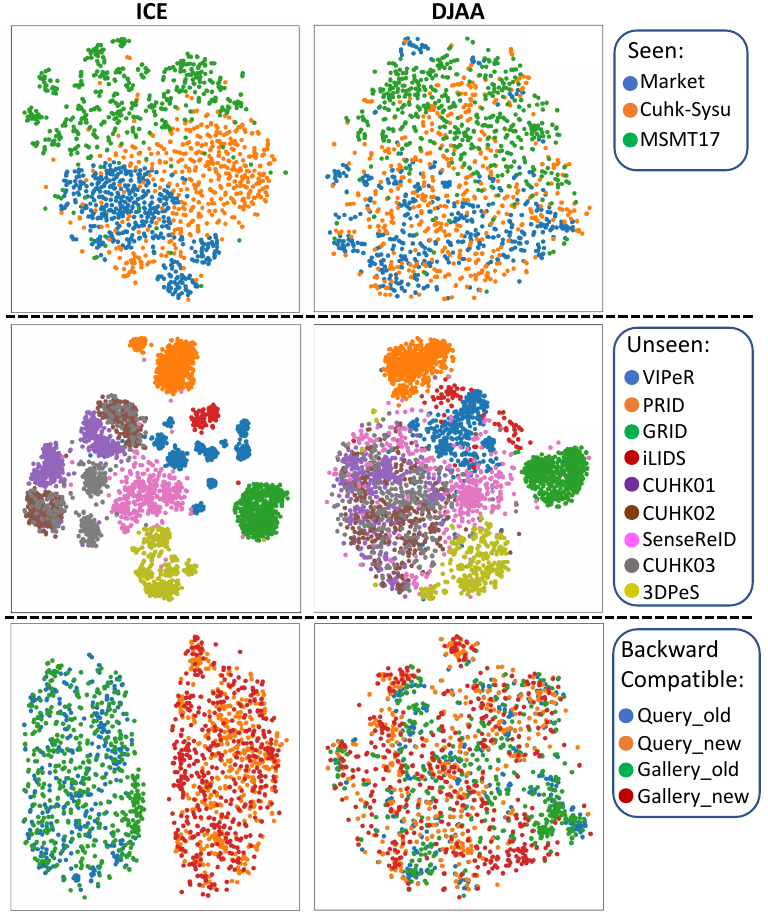}
   \caption{ReID feature space comparison~\cite{vandermaaten08a} of regular unsupervised ReID method ICE~\cite{Chen_2021_ICE} and unsupervised lifelong ReID method DJAA on three scenarios. 1) \textbf{Seen domain non-forgetting ability}: DJAA preserves previously acquired knowledge, which reduces the gap in the feature space between seen domains. 2) \textbf{Unseen domain generalization ability}: DJAA accumulates domain-shared features that reduce the gap between unseen domains. 3) \textbf{Backward-compatible ability}: With DJAA, updated query (Query\_new) and gallery (Gallery\_new) representations remain in the same feature space with the previously extracted representations (Query\_old, Gallery\_old). Query\_old and Gallery\_old are Market1501 representations extracted after one adaptation step, while Query\_new and Gallery\_new are Market1501 representations extracted after three adaptation steps.
   }
\label{fig:figure1}
\end{figure}

\IEEEPARstart{P}{erson} re-identification (ReID)~\cite{Ye_2021_reidsurvey} targets at matching a person of interest across non-overlapping cameras. Although significant improvement has been witnessed in both supervised~\cite{zheng2019joint,He_2021_ICCV} and unsupervised domain adaptive~\cite{zhong2020learning,Chen_2021_ICE} person ReID, traditional methods only consider the performance of a single fixed target domain. In the single target domain scenario, people usually assume that all training data are available before training a ReID model. However, a real-world video monitoring system can record new data every day and from new locations, when new cameras are installed into an existing system. 
When a model needs to be frequently updated with new data, regular unsupervised person ReID methods can lead to three problems: 1) Once adapted to a new domain, a model is prone to lose the acquired knowledge of previous seen domains. 2) A model can hardly learn domain-shared features and generalize to unseen domains. 3) An adapted retrieval model usually shows low backward-compatible ability.

Firstly, as the weather and season are repetitive, losing previous knowledge usually results in a less robust ReID system.
To tackle the forgetting problem, lifelong person ReID~\cite{pu_cvpr2021,Wu2021GeneralisingWF} has been recently proposed to incrementally accumulate domain knowledge from several seen datasets. Lifelong person ReID is closely related to incremental (or continuous) learning~\cite{van2022three}. There are three fundamental scenarios for incremental learning, including class-incremental learning, domain-incremental learning, and task-incremental learning. Facing non-stationary streams of data, lifelong person ReID is supposed to simultaneously learn incrementally added new classes and new domain knowledge, which can be defined as a joint class-incremental and domain-incremental learning task. However, previous supervised lifelong person ReID~\cite{pu_cvpr2021,Wu2021GeneralisingWF} relies on human annotations for cross-domain fine-tuning. Replacing supervised cross-domain fine-tuning with unsupervised domain adaptation can substantially enhance the flexibility of a lifelong person ReID algorithm in real-world deployments.


Secondly, to enhance the model generalization ability on unseen domain, domain generalization methods~\cite{Song2019GeneralizablePR} jointly train a model on several domains to learn domain-shared features. Regular domain generalization requires that several large-scale domain data are available before training a model. Lifelong person ReID can be an alternative for learning a generalizable model, which does not require collecting all data in advance. By regularizing the model consistency between old and new domains, we gradually accumulate domain-shared information into a unified model. With domain-shared features, an incrementally trained model~\cite{Wu2021GeneralisingWF} has proven to be generalizable on unseen domains.


Thirdly, after a domain adaptation step, an updated person ReID model also faces backward-compatiblility~\cite{shen2020towards,wan2022continual} problem. To obtain optimal performance on new data, a lifelong adaptive person ReID system needs to be frequently updated. During the lifelong adaptation, feature representations of previous gallery images need to be re-extracted to maintain a consistent feature space for the pairwise distance calculation. However, re-extracting features after each adaptation step can be time-consuming. An optimal setting for lifelong person ReID system is that an updated model only extracts feature representations from queries and incoming gallery images, while keeping previous gallery image representations unchanged. How to improve the feature consistency between a model and its updated versions remains a key problem for building an efficient system, which is neglected by previous lifelong person ReID methods. We show that anti-forgetting and backward-compatible ability can be unified into the same question for lifelong person ReID, \ie, preserving the representation similarity relationship.

In this paper, we propose an unsupervised lifelong person ReID method to simultaneously address the forgetting, generalization ability, and backward-compatibility problems. Unsupervised lifelong person ReID is a challenging yet practical problem. Compared with generic lifelong domain adaptation~\cite{tang2021gradient,rostami2021lifelong}, unsupervised lifelong person ReID faces more technical problems. On the one hand, lifelong domain adaptation usually considers fixed classes across different styles, while unsupervised lifelong person ReID has to deal with non-overlapping identities across domains. On the other hand, unsupervised lifelong ReID aims at learning fine-grained human features, which are usually more sensitive to domain changes than generic object features. Our objective is to train a robust person ReID model that is discriminative to each seen domain while being generalizable to unseen domains. 

Inter-instance Contrastive Encoding (ICE)~\cite{Chen_2021_ICE} formulates image-to-prototype and image-to-image contrastive learning to enhance unsupervised representation learning for person ReID. The proposed dual-level contrastive learning has proven to be effective in enhancing intra-cluster compactness and inter-cluster separability. ICE shows remarkable performance in unsupervised person ReID. However, when adapted to a new domain, ICE rapidly loses previously acquired knowledge from the source domain. In this paper, we extend ICE to unsupervised lifelong person ReID to enhance the anti-forgetting, generalization and backward-compatible ability during the multi-step adaptation, as shown in Fig.~\ref{fig:figure1}.
We incorporate pseudo-label based contrastive learning and rehearsal-based incremental learning into a Dual-level Joint Adaptation and Anti-forgetting (DJAA) framework, which enhances anti-forgetting, generalization and backward-compatible ability at the same time. 


Our proposed DJAA consists of an adaptation module and a rehearsal module. In the adaptation module, we first use a clustering algorithm to assign pseudo labels to unlabeled data. Based on the pseudo labels, we use an image-to-prototype contrastive loss to make images with the same pseudo-label converge to a unique cluster prototype. We also use an image-to-image contrastive loss to further reduce the distance between hard positives. In the rehearsal module, a small number of old domain samples and the corresponding cluster prototypes are selected and stored in a long-term memory buffer. While adapting a model to a new domain, regularizing the image-to-image and image-to-prototype similarity relationship helps to prevent forgetting previous knowledge. Given a frozen old domain model and a trainable model, we set a consistency regularization condition: the image-to-image and image-to-prototype similarity calculated by the frozen old domain model and the current domain model should be consistent during the adaptation. Based on this condition, we regularize the representation similarity relationship with both image-level and cluster-level similarity between the frozen and the trainable models, so that the trainable current domain model can be updated in a way that suits old knowledge.

The main contributions of our work are four folds. 
1) We comprehensively investigate the forgetting, generalization, and backward-compatible problems in domain adaptive person ReID. Our study shows that, those three challenges could be jointly addressed in a Dual-level Joint Adaptation and Anti-forgetting (DJAA) framework. To the best of our knowledge, this is an original person ReID work addressing those three challenges within a unified framework.
2) We propose an adaptation module that combines both cluster-level and instance-level contrastive losses to learn new domain features.
3) We propose a rehearsal module to retain previously-acquired knowledge during the domain adaptation. By introducing data augmentation and domain gap perturbations, we regularize the representation relationship at both cluster and instance levels.
4) Extensive experiments with various setups have been conducted to validate the effectiveness of our proposed method. DJAA shows remarkable non-forgetting, generalization, and backward-compatible ability on mainstream person ReID datasets.  

\section{Related Work}
\begin{table}
\centering
\caption{Comparison of supervised (S), unsupervised domain adaptation (UDA), domain generalization (DG), supervised lifelong (SL) and unsupervised lifelong (UL) ReID settings. `SD', `TD' and `UD' respectively refer to source domain, target domain and unseen domains.}
\scalebox{0.9}{
\begin{tabular}
{ccccc}
\hline
Setting& Domain & Train & Label & Test \\
\hline
S & one & TD & TD& TD\\
UDA &two &SD\&TD&SD&TD\\
DG &multi&all SD&SD&UD\\
SL&multi&one by one&SD&SD\&UD\\
UL&multi&one by one&None&SD\&UD\\
\hline
\end{tabular}}

\label{tab:ReID settings}
\end{table}
\subsection{Person ReID} Depending on the number of training/test domains and availability of human annotation, recent person ReID research is conducted under different settings, as shown in Table~\ref{tab:ReID settings}. As the most studied setting, supervised person ReID~\cite{He_2021_ICCV,lee2022negative,zhou2023adaptive,zhang2023pha} has shown impressive performance on large-scale datasets thanks to deep neural networks and human annotation. However, as a fine-grained retrieval task, a ReID model trained on one domain can hardly generalize to other domains. Unsupervised domain adaptation methods~\cite{liu2019adaptive,zheng2021exploiting,zheng2021group,ge2020self,Chen_2021_ICE,cho2022part} are proposed to adjust a ReID model to a target domain with unlabeled target domain images. To maximize the model generalization ability on unseen data, domain generalization ReID~\cite{Song2019GeneralizablePR,Jin_2020_CVPR,dai2021generalizable,zhang2022adaptive,jiao2022dynamically,xu2022mimic} is proposed to jointly train multiple labeled domains, in order to learn a generalizable model that can extract domain-invariant features. The above-mentioned settings simply assume that all training data is available before training. However, in most real-world cases, it is hard to prepare enough diversified data to directly train a generalizable model. Instead, new domain data can be recorded when time and season change or a new camera is installed. Supervised lifelong person ReID~\cite{pu_cvpr2021,Wu2021GeneralisingWF,lu2022augmented,ge2022lifelong,yu2023lifelong} is thus proposed to learn incrementally added data without forgetting previous knowledge. 
LSTKC~\cite{xu2024lstkc} introduces a relation matrix-based erroneous knowledge filtering and rectification mechanism to distill correct knowledge for supervised lifelong person ReID. However, continuously annotating new domains can be a cumbersome task for ReID system administrators. CLUDA~\cite{huang2022lifelong} combines meta learning and knowledge distillation to address the forgetting problem for Unsupervised lifelong person ReID. However, LSTKC~\cite{xu2024lstkc} and CLUDA~\cite{huang2022lifelong} have not considered the backward-compatible problem for lifelong person ReID. In addition, LSTKC and CLUDA neglect informative cluster prototypes that help to regularize the similarity relationship during adaptation. 

\subsection{Contrastive learning} The main idea of contrastive learning is to maximize the representation similarity between a positive pair composed of differently augmented views of a same image, so that a model can be invariant to view differences. While attracting a positive pair, some contrastive methods also consider other images as negatives and push away negatives stored in a memory bank~\cite{Wu2018UnsupervisedFL,He_2020_CVPR} or in a large mini-batch~\cite{chen2020simple}. 
Contrastive methods show great performance in unsupervised representation learning, which makes it the main approach in unsupervised person ReID. Based on clustering generated pseudo-labels, state-of-the-art unsupervised person ReID methods build positive pairs with cluster centroids~\cite{ge2020self}, camera-aware centroids~\cite{Wang2021camawareproxies} and generated positive views~\cite{Chen_2021_CVPR,chen2022learning}. ICE~\cite{Chen_2021_ICE} combines image-to-prototype and image-to-image contrastive losses to reach the maximal agreement between positive images and cluster prototypes. 
However, these contrastive methods only consider single-domain view agreements, which suffer from catastrophic forgetting in multi-domain learning. 

\subsection{Incremental learning} Incremental (also called continuous or lifelong) learning aims at learning new classes, domains or tasks without forgetting previously acquired knowledge. Previous methods can be roughly categorized into three directions, \ie, architecture-based, regularization-based and rehearsal-based methods.  Architecture-based methods combine task-specific parameters to build a whole network. These methods progressively extend network structure when new tasks are added into an existing model~\cite{rusu2016progressive, yoon2018lifelong, mallya2018packnet}. 
Regularization-based methods consist in regularizing model updates on new data in a way that does not contradict the old knowledge. A common approach is to freeze the old model as a teacher for previous knowledge distillation~\cite{Li2018LearningWF,douillard2021plop,shang2023incrementer}. 
Rehearsal-based (also called recall or replay) methods address the forgetting problem by storing a small number of old image samples~\cite{Rebuffi2017iCaRLIC,Castro2018End,cha2021co2l,luo2023class} or a generative model~\cite{choi2021dual,van2020brain}. The stored old data or generated data are used to remind the model of previous knowledge during incremental training. In addition to the above-mentioned supervised incremental methods, several attempts have been made in unsupervised lifelong adaptation, such as setting gradient regularization~\cite{tang2021gradient} in contrastive learning and consolidating the internal distribution~\cite{rostami2021lifelong}. CoTTA~\cite{wang2022continual} proposes to use weight-averaged and augmentation-averaged pseudo-labels for test-time adaptation. Differently, our method is designed for lifelong domain adaptation, where we regularize both cluster and instance-level relationship with data augmentation and domain gap perturbations in the training.
Moreover, general lifelong adaptation~\cite{tang2021gradient,rostami2021lifelong,wang2022continual} has identical classes across different domains, which is not suitable for lifelong ReID that has to learn fine-grained identity representations from non-overlapping classes across domains. 

\subsection{Backward Compatible Learning}
Backward-compatible learning aims at enhancing the backward consistency of feature representations, so that previously extracted representations can be comparable with newly extracted representations in retrieval. 
BCT~\cite{shen2020towards} employs a cross-entropy distillation loss between old and new classifiers to constrain new representations. Neighborhood Consensus Contrastive Learning (NCCL)~\cite{wu2022neighborhood} proposes a neighborhood consensus supervised contrastive learning method to constrain new representations at the sub-cluster level. AdvBCT~\cite{pan2023boundary} uses adversarial learning to minimize the distributional distance between old and new encoders. 
However, the above-mentioned backward-compatible learning usually focuses on single-step learning models. In a long session learning scenario, more adaptation steps make it easier to lose knowledge of beginning steps. 
To handle the multi-step learning scenario, Wan \etal ~\cite{wan2022continual} introduce a continual learner for visual search (CVS), which effectively improves the backward compatibility in the class-incremental task. 
CVS only considers class-incremental setting within a single domain, which is sub-optimal for the domain-incremental scenario. 
Oh \etal~\cite{oh2024lifelong} propose a part-assisted knowledge consolidation method that leverages both local and global features to enhance the backward compatibility in lifelong person ReID. Instead of using local features, our framework leverages informative cluster prototypes and instances to enhance the backward compatibility during domain-incremental learning.

\subsection{Difference with previous methods}
In this paper, we explore the possibility of jointly addressing the forgetting, generalization and backward-compatible problems existing in regular unsupervised person ReID methods. Although our method is based on contrastive learning, incremental learning and backward-compatible learning, there are major differences between our method and previous methods. 
Compared with previous contrastive learning that mainly considers single-domain performance, we propose to further reach maximal image-to-prototype and image-to-image similarity consistency across different domains for anti-forgetting. 
Compared with previous lifelong adaptation methods, our proposed unsupervised lifelong ReID method is able to accumulate fine-grained identity information from non-overlapping classes in a domain-incremental scenario. Moreover, our method uses a dual-level relationship regularization along with contrastive learning to better mitigate the forgetting problem. 
Previous backward-compatible learning methods, such as CVS~\cite{wan2022continual}, only consider class-incremental settings within a single domain, which is sub-optimal for the domain-incremental scenario. As incrementally added person ReID domains bring in both class and domain gaps, it is more difficult to retain backward consistency in lifelong ReID. Our proposed method regularizes the representation consistency at both cluster and instance levels, which enhances the backward compatibility during domain-incremental learning.

\section{Methodology}

\begin{figure*}
\centering
   \includegraphics[width=1\linewidth]{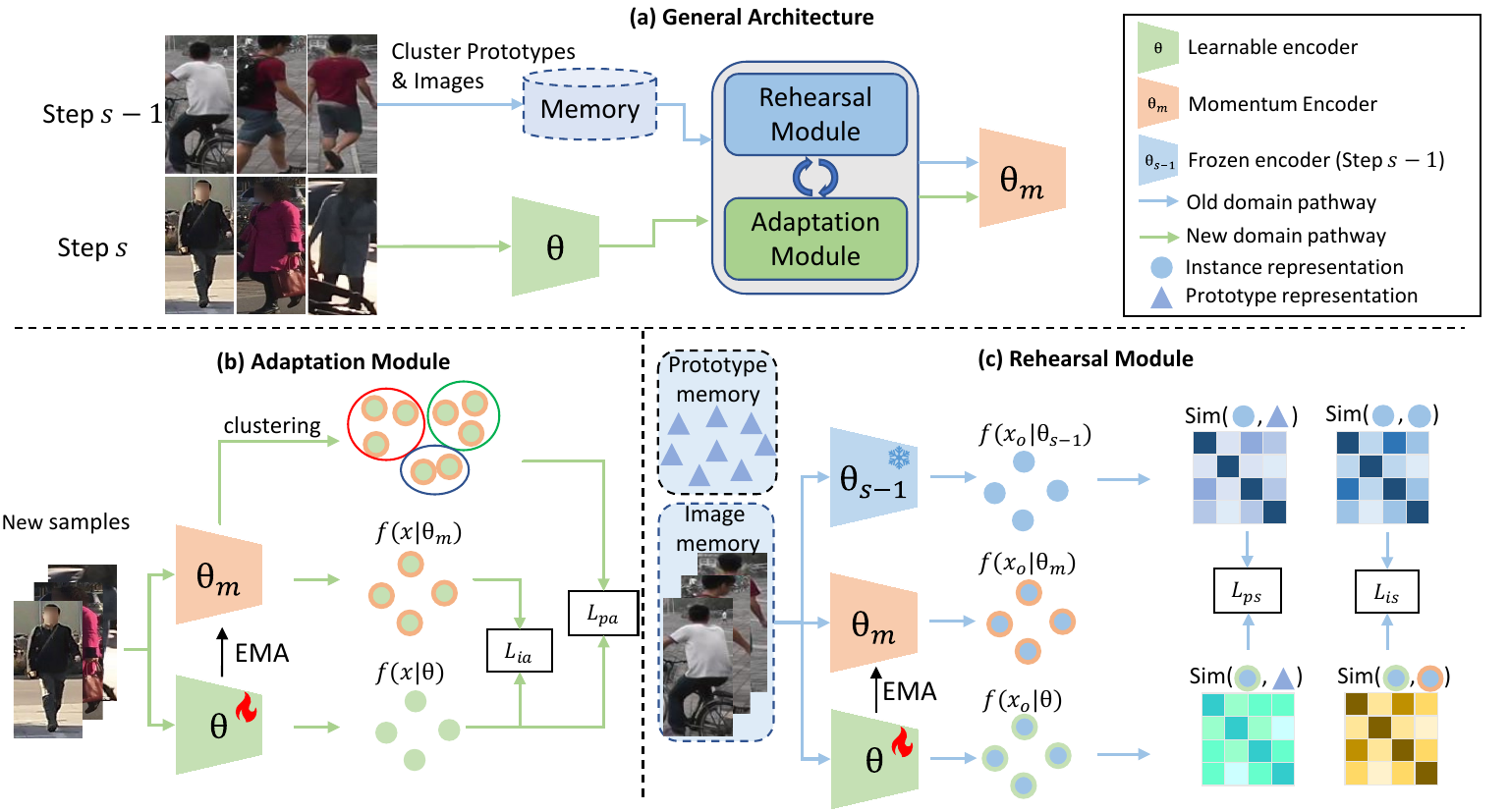}
   \caption{\textbf{(a) General architecture of DJAA}: our proposed framework is composed of an adaptation module and a rehearsal module. A memory buffer stores a small number of images and corresponding cluster prototypes from previous step $s-1$, which are trained jointly with data from step $s$. The two modules work collaboratively to achieve the adaptation without forgetting. \textbf{(b) Adaptation Module}: This module follows the new domain pathway (\textcolor{green}{$\rightarrow$}) to learn new domain knowledge with dual-level contrastive adaptation losses $\mathcal{L}_{pa}$ and $\mathcal{L}_{ia}$. \textbf{(c) Rehearsal Module}: This module follows the old domain rehearsal pathway(\textcolor{blue}{$\rightarrow$}) to rehearse old domain knowledge with dual-level similarity consistency losses $\mathcal{L}_{ps}$ and $\mathcal{L}_{is}$.}
\label{fig:figure2}
\end{figure*}

\subsection{Overview}
Given a stream of $N$ person ReID datasets, unsupervised lifelong person ReID aims at learning a generalizable model via sequential unsupervised learning on the training set of each domain $D_{1}\to ... \to D_{s-1}\to D_{s} ... \to D_{N}$. After the whole pipeline, the adapted model is tested respectively on the test set of each seen domain $D_{1}, ..., D_{N}$, as well as on multiple unseen domains. 

We use $\theta_{new}$ and $\theta_{old}$ to respectively represent the adapted and old models. Incremental learning aims at retaining previous knowledge while acquiring new knowledge. Given a triplet of an anchor $x_{i}$, a positive $x_{p}$ and a negative $x_{n}$, we have initially $d(f(x_{i}^{}|\theta_{old}),f(x_{p}^{}|\theta_{old})) < d(f(x_{i}^{}|\theta_{old}),f(x_{n}^{}|\theta_{old}))$ with the old model. After incremental adaptation, the distance between the anchor $f(x_{i}^{}|\theta_{new})$ and the positive $f(x_{p}^{}|\theta_{new})$ should remain smaller than that between the anchor and the negative $f(x_{n}^{}|\theta_{new})$. An incremental learning criterion can be defined as:
\begin{equation}
\begin{aligned}
&d(f(x_{i}^{}|\theta_{new}),f(x_{p}^{}|\theta_{new})) < d(f(x_{i}^{}|\theta_{new}),f(x_{n}^{}|\theta_{new})), \\
&\forall (i,p,n) \in \{ (i,p,n): y_{i} = y_{p} \ne y_{n}\},
\end{aligned}
\label{equ:Lifelong adaptation criterion}
\end{equation}
where $d(\cdot,\cdot)$ denotes the distance between two representations. 

Backward-compatible learning also aims at retaining previous knowledge while acquiring new knowledge. Different to incremental learning, backward-compatible learning focuses more on the compatibility between new query representations and old gallery representations. The distance between a new query $f(x_{i}^{}|\theta_{new})$ and a stored gallery positive $f(x_{p}^{}|\theta_{old})$ should be smaller than that between the query $f(x_{i}^{}|\theta_{new})$ and a stored gallery negative $f(x_{n}^{}|\theta_{old})$.
A backward-compatible criterion can be defined as:
\begin{equation}
\begin{aligned}
&d(f(x_{i}^{}|\theta_{new}),f(x_{p}^{}|\theta_{old})) < d(f(x_{i}^{}|\theta_{new}),f(x_{n}^{}|\theta_{old})), \\
&\forall (i,p,n) \in \{ (i,p,n): y_{i} = y_{p} \ne y_{n}\},
\end{aligned}
\label{equ:Backward-compatible criterion}
\end{equation}

In this paper, we show that the incremental learning criterion Eq.~(\ref{equ:Lifelong adaptation criterion}) and the backward-compatible criterion Eq.~(\ref{equ:Backward-compatible criterion}) can be both satisfied with a unified framework. 
We present a Dual-level Joint Adaptation and Anti-forgetting (DJAA) method for unsupervised lifelong person ReID. The general architecture of DJAA is illustrated in Fig.~\ref{fig:figure2} (a). To mitigate the forgetting, we build a hybrid memory buffer that stores a small number of informative images and corresponding cluster prototypes for rehearsing old knowledge. The proposed DJAA consists of an adaptation module and a rehearsal module, which work collaboratively to achieve the unsupervised lifelong adaptation without forgetting. For step $s$, we jointly train new samples on domain $D_{s}$ and old samples stored in the hybrid memory buffer to adapt the encoder $\theta_{s-1}$ to become the encoder $\theta_{s}$. 

The Adaptation Module of DJAA is shown in Fig.~\ref{fig:figure2} (b). We use the pseudo label based contrastive learning method ICE~\cite{Chen_2021_ICE} as a baseline, which leverages both an online encoder and a momentum encoder. In DJAA, the momentum encoder serves as a knowledge collector that gradually accumulates knowledge of each seen domain. During the step $s$, the momentum encoder (weights noted as $\theta_m$) accumulates new knowledge with exponential moving averaged weights of the online encoder (weights noted as $\theta_{}$):
\begin{equation}
\theta_m^t =\alpha\theta_m^{t-1}+(1-\alpha)\theta^t_{},
\label{equ:ema}
\end{equation}
where the hyper-parameter $\alpha$ controls the speed of the knowledge accumulation. $t$ and $t-1$ refer respectively to the current and last iterations. We extract all image representations with the stable momentum encoder and generate corresponding pseudo labels with a density-based clustering algorithm DBSCAN~\cite{Ester1996ADA}. Based on the clustered pseudo labels, we build cluster prototypes for prototype contrastive adaptation loss $\mathcal{L}_{pa}$ and image contrastive adaptation loss $\mathcal{L}_{ia}$ (described in Section \ref{sec:Adaptation Module}) on the new domain $D_{s}$. 

The Rehearsal Module of DJAA is shown in Fig.~\ref{fig:figure2} (c). Before adaptation,
we freeze the momentum encoder from the last step $\theta_{s-1}$ as an old knowledge expert model. We set the consistency regularization condition: the image-to-image and image-to-prototype similarity encoded by the frozen old domain model and the trainable model should be consistent during the adaptation. Based on the condition, 
we set an image-to-prototype similarity consistency loss $\mathcal{L}_{ps}$ and an image-to-image similarity consistency loss $\mathcal{L}_{is}$ (described in Section~\ref{sec:Rehearsal Module}) to regularize the model updates during the lifelong adaptation.

To train our proposed framework, we combine the above-mentioned four losses into an overall unsupervised lifelong loss:
\begin{equation}
\mathcal{L}_{overall} = \mathcal{L}_{pa}+\lambda_{ia}\mathcal{L}_{ia}+\lambda_{ps}\mathcal{L}_{ps} + \lambda_{is}\mathcal{L}_{is}.
\label{equ:overall}
\end{equation}


\subsection{Adaptation Module}
\label{sec:Adaptation Module}
Our adaptation module contains a prototype-level contrastive adaptation loss and an instance-level contrastive adaptation loss. Combining these two adaptation losses permits adapting a model to a new target domain, while reducing intra-cluster variance for better knowledge accumulation. 

\subsubsection{Prototype-level Contrastive Adaptation Loss}
Given an encoded query $q$ and a set of encoded samples $K$ = \{$k_0$, $k_1$, $k_2$, ...\}, we use $k_{+}$ to denote the positive match of the query $q$. We define a softmax cosine similarity function:
\begin{equation}
S(q,K,\tau) = \frac{\exp{(q \cdot k_{+}/\tau)}}{\sum\nolimits_{i=1}^{K}\exp{(q \cdot k_{i}/\tau)}},
\label{equa:softmax cosine}
\end{equation}
where $q \cdot k$ is the cosine similarity between $q$ and $k$. $\tau$ is a temperature parameter that controls the similarity scale.

Inside an unsupervised lifelong ReID pipeline, our model incrementally learns new knowledge on each domain. For step $s$, $D_s=\{x_{1}, ..., x_{n}\}$ where $n$ is the number of current domain unlabeled images. We first use the momentum encoder $\theta_{m}$ to extract image representations and calculate re-ranking~\cite{zhong2017re} based similarity between each image pair. Based on the pair-wise similarity, the DBSCAN clustering algorithm is utilized to assign pseudo labels $\{y_{1}, ..., y_{n}\}$ to unlabeled images $\{x_{1}, ..., x_{n}\}$. Given a current domain image $x_{i}$, $f(x_{i}^{}|\theta)$ and $f(x_{i}^{}|\theta_m)$ denote respectively the online and the momentum representations. The prototype of a cluster $j$ is defined as the averaged momentum representations of all the samples with a same pseudo-label $y_{j}$:
\begin{equation}
   p^{}_{j} = \frac{1}{n_{j}}\sum_{x_{i}^{} \in y_{j}} f(x_{i}^{}|\theta_m),
\label{equa:cluster centroid}
\end{equation}
where $n_{j}$ is the number of images in the cluster $j$.

We use $P$ to denote all the cluster prototypes in the current domain. A cluster prototype adaptation loss maximizes the similarity between a sample $x_{i}$ and the positive prototype in $P$, which can be defined as:
\begin{equation}
\mathcal{L}_{pa} = -\log S(f(x_{i}^{}|\theta), P, \tau_{pa})
\label{equa:cluster contrast}
\end{equation}
where $\tau_{pa}$ is a temperature hyper-parameter.

\subsubsection{Instance-level Contrastive Adaptation Loss}
The prototype-level adaptation loss $\mathcal{L}_{pa}$ makes samples from the same cluster converge to a common prototype and push them away from other clusters. However, images belonging to the same class can be easily affected by noisy factors, such as illumination and view-point, leading to high intra-class variance. When the adaptation module is only trained with $\mathcal{L}_{pa}$, all the image samples converge at a same pace. Consequently, representations of hard positive samples remain far away after optimization. We use an identity-aware sampler to construct mini-batches. A mini-batch $X$ is composed of $n_p$ identities, where each identity has $n_k$ positive instances. Thus, the batch-size is $n_{bs}=n_p\times n_k$. Given an anchor instance $f(x_{i}^{}|\theta)$, we sample the hardest positive momentum representation $f(x_{j}^{}|\theta_m)$ that has the lowest cosine similarity with $f(x_{i}^{}|\theta)$. To improve the intra-class compactness, we formulate an instance contrastive adaptation loss:
\begin{equation}
\mathcal{L}_{ia} = -\log S(f(x_{i}^{}|\theta), f(X_{}^{}|\theta_m), \tau_{ia})
\label{equa:instance contrast}
\end{equation}
where $f(X_{}^{}|\theta_m)$ denotes the mini-batch containing one hard positive $f(x_{j}^{}|\theta_m)$ and $(n_p-1)\times n_k$ negatives, and $\tau_{ia}$ is a temperature hyper-parameter. 

\begin{table}
\centering
\caption{Comparison of positive view selection methods on Market. }
\scalebox{1}{
\begin{tabular}{l|cc}
\hline
\multirow{2}{*}{Method} & \multicolumn{2}{c}{Market} \\ \cline{2-3}
\multicolumn{1}{c|}{} & \multicolumn{1}{c}{mAP} & \multicolumn{1}{c}{R1} \\ 
\hline
Top-1 hardest positive (ours)&\textbf{82.3}&\textbf{93.8}\\
Average of top-2 hard positives&\textbf{82.3}&93.4\\
Average of top-3 hard positives&81.8&93.1\\
\hline
\end{tabular}}
\label{table: topk hard}
\end{table}

\textbf{Remark.} 
In our instance-level contrastive adaptation loss, we select the hardest positive to maximally increase the similarity between an anchor and the hard positive. Another possibility is to use the average of top-k hard positives as the contrastive positive view. We compare the positive selection methods in Table~\ref{table: topk hard}. The results show that the hardest positive slightly outperforms the average of top-k hard positives. We thus choose to select the hardest positive in our method.

By reducing the intra-cluster variance between hard positive samples, the instance-level adaptation loss $\mathcal{L}_{ia}$ allows for purifying the new knowledge on $D_{i}$ before being accumulated into the model $\theta_{m}$. Person ReID is a cross-camera image retrieval task, where the individual camera style is the main factor that causes intra-cluster variance. As shown in ICE~\cite{Chen_2021_ICE}, using camera labels to reduce camera style variance can further mitigate the noise during the knowledge accumulation. 
A camera proxy $p_{jk}$ is defined as the averaged momentum representations of all the instances belonging to the cluster $j$ captured by camera $c_k$: 
\begin{equation}
p^{}_{jk} = \frac{1}{n_{jk}}\sum_{x_{i} \in y_{j} \cap x_{i} \in c_k}^{}  f(x_{i}^{}|\theta_m),
\label{equa:center_c}
\end{equation}
where $n_{jk}$ is the number of instances belonging to the cluster $j$ captured by camera $c_k$. We form a set of camera prototypes $P_{cam}$, which combines one positive cross-camera prototype and $n_{neg}$ nearest negative prototypes. A camera contrastive loss is the softmax log loss that minimizes the distance between an anchor $f(x_{i}^{}|\theta)$ and each of positive cross-camera prototypes:
\begin{equation} 
\mathcal{L}_{cam} = -\frac{1}{|\mathcal{P}|} \sum_{x_{i}\in y_{j} \cap x_{i} \notin c_k} \log S(f(x_{i}^{}|\theta), P_{cam}, \tau_{c}),
\label{equa:cameraloss}
\end{equation}
where $\tau_{c}$ is a cross-camera temperature hyper-parameter, which is set to 0.07 following \cite{Chen_2021_ICE}. 
$|\mathcal{P}|$ is the number of cross-camera positive prototypes. 
We provide extra results with the camera contrastive loss for knowledge accumulation in Section~\ref{sec: Experiments}.

\subsection{Rehearsal Module}
\label{sec:Rehearsal Module}
\begin{figure}[t]
\centering
   \includegraphics[width=1\linewidth]{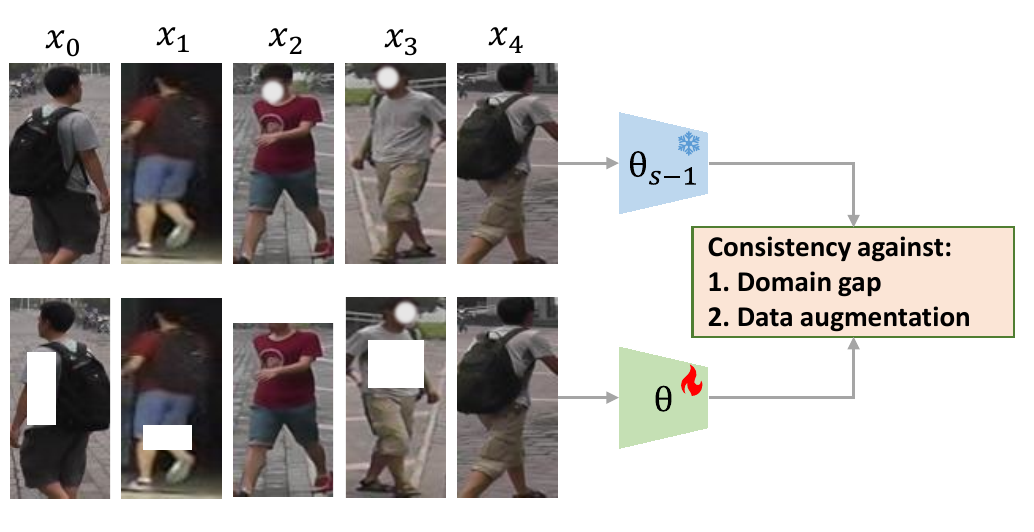}
   \caption{We introduce two kinds of perturbations, including domain gap from different encoders and data augmentation. Our target is to make the model invariant to domain gap and data augmentation perturbations. }
\label{fig:perturbations}
\end{figure}

In our rehearsal module, buffered samples and prototypes are utilized to prevent catastrophic forgetting during the domain adaptation. This module contains an Image-to-prototype Similarity Consistency loss and Image-to-image Similarity Consistency loss to maximally retain the old knowledge. ICE~\cite{Chen_2021_ICE} introduces a consistency regularization method to enhance a model's invariance to data augmentation perturbations. In this paper, we further consider the encoder style as a kind of perturbation. As shown in Figure~\ref{fig:perturbations}, we enhance the similarity consistency between differently augmented views encoded by current domain encoder $\theta$ and last step encoder $\theta_{s-1}$. In this way, our model can be robust to both domain style and data augmentation changes. 

\subsubsection{Image-to-prototype Similarity Consistency loss}
\label{sec:Image-to-prototype Similarity Consistency loss}
Technically, person ReID is a representation similarity ranking problem, in which the objective is to have high similarity scores between positive pairs and low similarity scores between negative pairs. However, when a model is adapted into a new domain, the similarity relationship between old domain samples could be affected by the new domain knowledge.
As a cluster prototype (the averaged representation of all cluster samples) contains generic information of a cluster, the prototype memory enables the current model to have access to generic old domain cluster information without storing all the images. For the memory buffer of $n_{mem}$ buffered images and prototypes, we first use the buffered cluster prototypes $P^{o}=\{p^{o}_1, ..., p^{o}_{n_{mem}}\}$ as anchors to rehearse the old domain knowledge. As the similarity relationship between images and prototype should be consistent before and after a domain adaptation step, we propose an image-to-prototype similarity loss to ensure the upcoming new knowledge would not change the image-to-prototype similarity.

As the frozen old model $\theta_{s-1}$ from the last domain can be regarded as an expert on the old domain, we calculate the cosine similarity between representations encoded by $\theta_{s-1}$ and prototypes as a reference to regularize the online model $\theta_{}$.
Given a mini-batch of old images $\{x_{1}^{o}, ..., x_{n_{bs}}^{o}\}$ randomly sampled from the memory buffer where $n_{bs}$ is the batch size, the online image-to-prototype similarity between a buffered image $x_{i}^{o}$ and buffered prototypes $P^{o}_{}$ is defined as $S(f(x_{i}^{o}|\theta_{}), P^{o}, \tau_{ps})$.
For the same mini-batch, we calculate the image-to-prototype similarity with the frozen model $\theta_{s-1}$ as the reference similarity. The reference similarity between the same image-prototype pair is defined as $S(f(x_{i}^{o}|\theta_{s-1}), P^{o}, \tau_{ps})$.

We formulate an image-to-prototype similarity consistency loss with a Kullback-Leibler (KL) Divergence between the two similarity distributions: 
\begin{equation}
\begin{aligned}
   \mathcal{L}_{ps} = \mathcal{D}_{KL}(&S(f(x_{i}^{o}|\theta_{}), P^{o}, \tau_{ps})||\\
   &S(f(x_{i}^{o}|\theta_{s-1}), P^{o}, \tau_{ps})).
\end{aligned}
\label{equa:ps}
\end{equation}
The buffered prototypes can be regarded as anchors of previously-acquired knowledge. We expect that the new domain knowledge does not change the distance relationship between an image and the prototypes, so that we can retain the previously-acquired knowledge for anti-forgetting adaptation. KL divergence measures the information loss when one distribution is used to approximate another. By minimizing $\mathcal{L}_{ps}$, we encourage the image-to-prototype similarity distribution $S(f(x_{i}^{o}|\theta_{}), P^{o}, \tau_{ps})$ calculated with current domain knowledge to be consistent with that calculated with old domain distribution $S(f(x_{i}^{o}|\theta_{s-1}), P^{o}, \tau_{ps})$.

\subsubsection{Image-to-image Similarity Consistency loss}
\label{sec:Image-to-image Similarity Consistency loss}
Cluster prototypes contain general cluster information, for example, the most salient feature shared by different views of a person. Differently, image instances contain more detailed information in a specific view, which is complementary to prototype-level general information. 
As the similarity relationship between the same images should be consistent before and after a domain adaptation step, we propose an image-to-image similarity loss that regularizes the similarity relationship updates in a way that does not contradict the old knowledge.
Similar to the image-to-prototype similarity, we also use the frozen old model $\theta_{s-1}$ from the last domain as an expert to calculate the reference similarity.
Given a mini-batch of old images $X^o= \{x_{1}^{o}, ..., x_{n_{bs}}^{o}\}$ where $n_{bs}$ is the batch size, the image-to-image similarity distribution can be calculated with a softmax function on the cosine similarity between each image pair in the mini-batch. The image-to-image similarity between an old image $x_{i}^{o}$ and a mini-batch $X^o$ is calculated with both the online encoder $\theta_{}$ and the momentum encoder $\theta_{m}$, \ie,  $S(f(x_{i}^{o}|\theta_{}), f(X^{o}_{}|\theta_{m}), \tau_{is})$.
For the same mini-batch, we calculate the image-to-image similarity distribution with the frozen old model $\theta_{s-1}$ as a reference for the constraint. The reference similarity between the same image $x_{i}^{o}$ and the mini-batch $X_{}^{o}$ is $S(f(x_{i}^{o}|\theta_{s-1}),  f(X^{o}_{}|\theta_{s-1})), \tau_{is})$.

We formulate an image-to-image similarity constraint loss with the KL Divergence between the two distributions:
\begin{equation}
\begin{aligned}
    \mathcal{L}_{is} = \mathcal{D}_{KL}(&S(f(x_{i}^{o}|\theta_{}), f(X^{o}_{}|\theta_{m}), \tau_{is})||\\
   &S(f(x_{i}^{o}|\theta_{s-1}),  f(X^{o}_{}|\theta_{s-1})), \tau_{is})).
\end{aligned}
\label{equa:is}
\end{equation}
By minimizing $\mathcal{L}_{is}$, we encourage the similarity relationship calculated with current domain knowledge $\theta_{}$ to be consistent with that calculated with old domain knowledge $\theta_{s-1}$. 

\textbf{Remark.} 
As shown in Fig.\ref{fig:perturbations}, we introduce both domain gap and data augmentation perturbations. The domain gap perturbations are introduced by using online and frozen encoders. The data augmentation perturbations are introduced by using two data augmentation settings on same images. Inspired by consistency regularization from semi-supervised learning~\cite{sohn2020fixmatch}, we use weak data augmentation on reference similarity calculation and strong data augmentation on prediction similarity calculation. As data augmentation can mimic image perturbations, augmentation consistency regularization allows for enhancing the model robustness against perturbations in real deployments. 


\subsection{Memory Buffer Update}
\label{sec:Memory Buffer Update}

We store a small number of informative samples and cluster prototype representations in a hybrid memory buffer, which is updated at the end of each step. The total size of the hybrid memory buffer is set to $n_{mem}$ images and $n_{mem}$ prototypes. In our proposed DJAA framework, the default value of $n_{mem}$ is $512$. Suppose that $|P|$ is the number of current domain clusters and $|P^{o}|$ is the number of buffered cluster prototypes, we update the memory buffer with $n_{new}$ samples from the new domain and $n_{old}$ samples from the memory buffer:
\begin{equation}
n_{new} = \frac{|P|}{|P^{o}|+|P|} \times n_{mem},
\label{equa:update number new}
\end{equation}
\begin{equation}
n_{old} = \frac{|P^{o}|}{|P^{o}|+|P|} \times n_{mem}.
\label{equa:update number old}
\end{equation}

Previous methods~\cite{huang2022lifelong,ge2022lifelong} usually store several samples from randomly selected classes, which is sub-optimal for our proposed image-to-prototype similarity consistency loss. 
We propose a simple yet effective clustering-guided selection algorithm to select informative samples. As the cluster prototype is defined as the averaged representation, the prototype can cover more sample information if a cluster contains more samples. We thus use the number of samples to rank each cluster and preferentially select clusters that contain more samples. Once a cluster is selected, we proceed to calculate the cosine similarity between the cluster prototype and cluster samples. Under unsupervised setting, we argue that the sample with the highest similarity score is the most credible sample belonging to the cluster. To mitigate the pseudo-label noise and enhance the data diversity, we select 1 sample that is closest to the prototype in each selected cluster. We provide training and memory buffer updating details in Algorithm~\ref{algo:1}. We validate the effectiveness of our proposed memory buffer updating method in Section~\ref{Data selection for memory buffer update}.

\begin{algorithm}[t]
\caption{DJAA for unsupervised lifelong person ReID}
\label{algo:1}
\textbf{Input}: Unlabeled domains $D_{1}\to ... \to D_{s} \to... \to D_{N}$, a hybrid memory buffer of size $n_{mem}$, an online encoder $\theta$ and a momentum encoder $\theta_{m}$. \\
\textbf{Output}: Encoder $\theta_m$ after $N$-step adaptation. \\
\begin{algorithmic}[1] 
\FOR {$D_{s}=D_{1}$ to $D_{N}$}
    \STATE Get $\theta_m$ from the last step $D_{s-1}$. Freeze a copy of momentum encoder $\theta_{s-1} \gets \theta_m$. Initialize the online encoder with the momentum encoder $\theta \gets \theta_m$;
    \FOR {$epoch=1$ to $E_{max}$}
        \STATE Generate pseudo labels on $D_s$;
        \STATE Calculate cluster prototypes $P$ in Eq.~(\ref{equa:cluster centroid}) on $D_s$;
        \FOR {$iter=1$ to $I_{max}$}
            \STATE Sample a mini-batch from the current domain $D_s$\ for $\mathcal{L}_{pa}$ in Eq.~(\ref{equa:cluster contrast}) and $\mathcal{L}_{ia}$ in Eq.~(\ref{equa:instance contrast});
            \STATE Sample a mini-batch from the memory buffer for $\mathcal{L}_{ps}$ in Eq.~(\ref{equa:ps}) and $\mathcal{L}_{is}$ in Eq.~(\ref{equa:is});
            \STATE Train $\theta$ with $\mathcal{L}_{overall}$ in Eq.~(\ref{equ:overall});
            \STATE Update $\theta_m$ with Eq.~(\ref{equ:ema});
        \ENDFOR
    \ENDFOR
    \STATE Update the memory buffer with $n_{new}$ new samples and $n_{old}$ previous samples, following Eq.~(\ref{equa:update number new}) and Eq.~(\ref{equa:update number old});
    \STATE Store $\theta_m$ for next step $D_{s+1}$;
\ENDFOR
\end{algorithmic}
\end{algorithm}

\section{Experiments}
\label{sec: Experiments}
\subsection{Datasets and Evaluation Protocols}
We use 4 seen datasets for domain-incremental training and 10 unseen datasets for generalization ability evaluation, as shown in Table~\ref{tab:datasets}. 
\begin{table}[t]
\centering
\caption{Dataset statistics. Unseen domains are only used for testing.}
\scalebox{0.9}{
\begin{tabular}
{c|ccccc}
\hline
Type& Dataset  & \#train img & \#train id & \#test img & \#test id\\
\hline
\multirow{4}{*}{Seen} & PersonX~\cite{sun2019dissecting}&9840&410&35952&856\\
& Market~\cite{Zheng2015ScalablePR} & 12936 & 751& 19281 & 750\\
 & Cuhk-Sysu~\cite{Xiao2017JointDA} & 15088 & 5532& 8347 & 2900\\
 & MSMT17~\cite{wei2018person} & 32621 & 1041 & 93820 & 3060\\
\hline
\multirow{10}{*}{Unseen} & VIPeR~\cite{Gray2008ViewpointIP}&-&-&632&316\\
 & PRID~\cite{hirzer11}&-&-&749&649\\
 & GRID~\cite{Loy2009MulticameraAC}&-&-&1025&125\\
 & iLIDS~\cite{Zheng2009AssociatingGO}&-&-&120&60\\
 & CUHK01~\cite{Li2012HumanRW}&-&-&1944&486\\
 & CUHK02~\cite{Li2013LocallyAF}&-&-&956&239\\
 & SenseReID~\cite{Zhao2017SpindleNP}&-&-&4428&1718\\ 
 & CUHK03~\cite{Li2014DeepReIDDF}&-&-&1930&100\\
 & 3DPeS~\cite{3dpes}&-&-&512&96\\
 & MMP-Retrieval~\cite{huang2022lifelong}&-&-&28907&7\\
 
\hline
\end{tabular}}
\label{tab:datasets}
\end{table}

\textbf{Incremental training on seen datasets:} We set 2 incremental training pipelines on seen datasets, \ie, one-cycle full set benchmark and two-cycle subset benchmark. The one-cycle full-set benchmark targets at evaluating the effectiveness of handling imbalanced class numbers between different adaptation steps, while the two-cycle subset benchmark aims to mimic the season and weather cycle, which may re-appear after several adaptation steps. In the one-cycle full set benchmark, we do not use any supervised pre-training. The model is directly adapted to Market1501, CUHK-SYSU, and MSMT17. We further define two training orders, \ie, Market$\to$Cuhk-Sysu$\to$MSMT17 and MSMT17$\to$Market$\to$Cuhk-Sysu. In the two-cycle subset benchmark, following CLUDA~\cite{huang2022lifelong}, we first pre-train our model on PersonX in a supervised manner. Then, we adapt our pre-trained model to Market1501, CUHK-SYSU, and MSMT17 in an unsupervised manner. The adaptation steps are repeated twice, with each stage involving a subset of 350 identities. The two-cycle training order is defined as PersonX$\to$Market $\to$Cuhk-Sysu$\to$MSMT17$\to$Market$\to$Cuhk-Sysu$\to$MSMT17. Cumulative Matching Characteristics (CMC) at Rank1 accuracy and mean Average Precision (mAP) are used in our experiments. We also report the averaged $\Bar{s}Rank1$ and $\Bar{s}mAP$ on the seen domains and unseen domains. 

\textbf{Generalization ability on unseen datasets:} We use 10 person ReID datasets to maximally evaluate the model generalization ability on different unseen domains, including VIPeR, PRID, GRID, iLIDS, CUHK01, CUHK02, SenseReID, CUHK03, 3DPeS and MMP-Retrieval. These 10 datasets cover all the unseen domains that are considered in previous supervised lifelong ReID methods~\cite{Wu2021GeneralisingWF,pu_cvpr2021} and domain generalizable ReID methods~\cite{Song2019GeneralizablePR}. We use the traditional training/test split on CUHK03 dataset. Rank1 accuracy and mAP results are respectively reported on the test set of each unseen domain after the final step. 

\textbf{Backward-compatible evaluation:} To evaluate the backward compatibility of feature representations, we report retrieval performance between current query images and current gallery images (termed as self-test), and that between current query images and stored gallery images (termed as cross-test). 

\subsection{Implementation details}
\subsubsection{Training} Our method is implemented under Pytorch~\cite{PyTorch_NEURIPS2019} framework. The total training time with 4 Nvidia 1080Ti GPUs is around 6 hours. DJAA supports multiple hardware platforms and already supports training and deployment on Ascend 910B NPUs. We use an ImageNet \cite{Russakovsky2015ImageNetLS} pre-trained ResNet50 \cite{he2016deep} as our backbone network. For the strong data augmentation, we resize all images to 256$\times$128 and augment images with random horizontal flipping, cropping, Gaussian blurring and erasing \cite{Zhong2020RandomED}. For the weak data augmentation, we only resize images to 256$\times$128. 

At each step in the one-cycle full set training, we train our framework 30 epochs with 400 iterations per epoch using a Adam~\cite{Adam_optimization} optimizer with a weight decay rate of 0.0005. For the two-cycle subset evaluation, we follow CLUDA~\cite{huang2022lifelong} to train our model for 60 epochs without specified iterations. The learning rate is set to 0.00035 with a warm-up scheme in the first 10 epochs. No learning rate decay is used in the training. Pseudo labels on the current domain are updated on re-ranked Jaccard distance~\cite{zhong2017re} at the beginning of each epoch with a DBSCAN~\cite{Ester1996ADA}, in which the minimum cluster sample number is set to 4 and the distance threshold is set to 0.55. The momentum encoder is updated with a momentum hyper-parameter $\alpha=0.999$. To balance the model ability on old domains and the new domain, we separately take a mini-batch of current domain images and a mini-batch of buffered images of the same batch size $n_{bs}$, which is set to $n_{bs}=32$ in our experiments. Furthermore, we use a random identity sampler to construct mini-batches to handle the imbalanced images of different identities. Following the clustering setting on the current domain, the 32 current domain images are composed of 8 identities and 4 images per identity. Inside the mini-batch of buffered images, the 32 buffered images are composed of 32 identities and 1 image per identity. 

We use grid search to set the optimal temperature and balancing hyper-parameters in our proposed losses. Based on grid search results, we set the temperature hyper-parameters $\tau_{pa}=0.5$, $\tau_{ia}=0.1$, $\tau_{ps}=0.1$ and $\tau_{is}=0.2$. To make rehearsal and adaptation losses on the same scale, we set the balancing hyper-parameters $\lambda_{ia}=1$, $\lambda_{ps}=10$ and $\lambda_{is}=20$. For the memory buffer, we set $n_{mem}=512$, where the total identity number equals 512 and 1 image per cluster. After the whole training, only the momentum encoder is saved for inference.

\begin{table*}[t]
\centering
\caption{Seen-domain results (\%) of unsupervised domain adaptation (U) and unsupervised lifelong (UL) methods on one-cycle full set benchmark. The training order is Market$\to$Cuhk-Sysu$\to$MSMT17. * refers to using the camera loss~Eq.(\ref{equa:cameraloss}) to reduce intra-cluster variance. The best performance is marked in bold. }
\scalebox{0.9}{
\begin{tabular}{l|c|c|cc|cc|cc|cc}
\hline
\multirow{2}{*}{Method} & Memory & \multirow{2}{*}{Type} & \multicolumn{2}{c|}{Market} & \multicolumn{2}{c|}{Cuhk-Sysu} & \multicolumn{2}{c|}{MSMT17} & \multicolumn{2}{c}{Average} \\ \cline{4-11}
\multicolumn{1}{c|}{} &\multicolumn{1}{c|}{size}&\multicolumn{1}{c|}{}& \multicolumn{1}{c}{mAP} & \multicolumn{1}{c|}{Rank1} &\multicolumn{1}{c}{mAP} & \multicolumn{1}{c|}{Rank1} &\multicolumn{1}{c}{mAP} & \multicolumn{1}{c|}{Rank1} &\multicolumn{1}{c}{$\bar{s}_{mAP}$} & \multicolumn{1}{c}{$\bar{s}_{Rank1}$}  \\ 
\hline
ICE~\cite{Chen_2021_ICE}&0&U&29.0&60.4&72.5&76.3&21.8&49.0&41.1&61.9\\
CC~\cite{dai2022cluster}&0&U&31.0&58.9&74.6&77.3&25.7&51.8&43.8&62.7\\
PPRL~\cite{cho2022part}&0&U&29.5&58.6&75.6&79.5&\textbf{32.9}&\textbf{63.2}&46.0&67.1\\
LwF~\cite{Li2018LearningWF}&0&UL&27.5&59.0&70.5&74.7&20.3&48.6&39.5&60.8\\
iCaRL~\cite{Rebuffi2017iCaRLIC}&512&UL&37.4&67.6&79.5&81.9&19.9&45.4&45.5&65.0\\
C$o^2$L~\cite{cha2021co2l}&512&UL&35.3&62.0&78.3&80.8&24.2&50.7&46.0&64.5\\
CVS~\cite{wan2022continual}&512&UL&56.8&78.7&74.6&77.4&15.7&38.6&49.0&64.9\\
LSTKC~\cite{xu2024lstkc}&512&UL&48.8&74.9&77.2&79.7&20.6&47.6&48.8&67.4\\
DJAA (ours)&512&UL&60.2&82.5&\textbf{83.9}&\textbf{85.6}&20.7&46.7&54.9&71.6\\
DJAA* (ours)&512&UL&\textbf{65.2}&\textbf{86.3}&81.8&84.1&23.7&51.6&\textbf{56.9}&\textbf{74.0}\\
\hline
\end{tabular}}

\label{table:seen domains results}
\end{table*}

\begin{table*}[t]
\centering
\caption{Unseen-domain results (\%) of unsupervised (U), unsupervised lifelong (UL) and domain generalization (DG) methods on one-cycle full set benchmark. The training order is Market$\to$Cuhk-Sysu$\to$MSMT17. * refers to using the camera loss~Eq.(\ref{equa:cameraloss}) to reduce intra-cluster variance. The best unsupervised performance is marked in bold.}
\scalebox{0.9}{
\setlength{\tabcolsep}{3pt}
\begin{tabular}{l|c|c|cc|cc|cc|cc|cc|cc|cc|cc|cc|cc}
\hline
\multirow{2}{*}{Method} & Memory&\multirow{2}{*}{Type}& \multicolumn{2}{c|}{VIPeR} & \multicolumn{2}{c|}{PRID} & \multicolumn{2}{c|}{GRID} & \multicolumn{2}{c|}{iLIDS} & \multicolumn{2}{c|}{CUHK01}& \multicolumn{2}{c|}{CUHK02}& \multicolumn{2}{c|}{SenseReID}& \multicolumn{2}{c|}{CUHK03}& \multicolumn{2}{c|}{3DPeS}& \multicolumn{2}{c}{Average} \\ \cline{4-23}
\multicolumn{1}{c|}{}&\multicolumn{1}{c|}{Size}&\multicolumn{1}{c|}{}& \multicolumn{1}{c}{mAP} & \multicolumn{1}{c|}{R1} &\multicolumn{1}{c}{mAP} & \multicolumn{1}{c|}{R1} &\multicolumn{1}{c}{mAP} & \multicolumn{1}{c|}{R1} &\multicolumn{1}{c}{mAP} & \multicolumn{1}{c|}{R1} &\multicolumn{1}{c}{mAP} & \multicolumn{1}{c|}{R1} &\multicolumn{1}{c}{mAP} & \multicolumn{1}{c|}{R1} &\multicolumn{1}{c}{mAP} & \multicolumn{1}{c|}{R1} &\multicolumn{1}{c}{mAP} & \multicolumn{1}{c|}{R1} &\multicolumn{1}{c}{mAP} & \multicolumn{1}{c|}{R1} &\multicolumn{1}{c}{$\bar{s}_{mAP}$} & \multicolumn{1}{c}{$\bar{s}_{R1}$}  \\ 
\hline
ICE~\cite{Chen_2021_ICE}&0&U&35.7&25.9&39.0&29.0&20.6&14.4&71.4&61.7&60.6&60.0&48.2&45.8&33.6&27.9&17.3&29.7&48.5&55.4&41.7&38.9\\
CC~\cite{dai2022cluster}&0&U&43.3&32.6&41.6&28.0&21.1&15.2&79.5&73.3&66.3&66.0&57.2&57.3&37.9&30.6&25.6&24.0&54.4&59.6&47.4&43.0\\
PPRL~\cite{cho2022part}&0&U&40.5&30.1&43.8&33.0&15.0&9.6&74.3&63.3&66.3&65.9&54.5&53.3&38.4&31.2&22.2&20.3&50.0&62.4&45.0&41.0\\
LwF~\cite{Li2018LearningWF}&0&UL&41.0&29.7&40.1&30.0&19.7&13.6&74.9&66.7&62.4&61.9&52.8&51.9&33.9&27.3&20.0&34.1&52.2&58.6&44.1&41.5\\
iCaRL~\cite{Rebuffi2017iCaRLIC}&512&UL&48.1&38.0&44.5&34.0&29.4&20.0&82.1&76.7&66.9&66.0&58.3&55.2&42.1&35.5&26.2&41.3&57.1&63.6&50.5&47.8\\
C$o^2$L~\cite{cha2021co2l}&512&UL&43.8&32.0&50.7&41.0&30.4&21.6&\textbf{83.6}&\textbf{76.7}&69.5&69.3&58.1&54.2&39.6&31.6&25.3&41.6&55.4&62.7&50.7&47.8\\
CVS~\cite{wan2022continual}&512&UL&44.3&33.2&31.6&23.0&31.0&23.2&78.5&71.7&62.8&60.8&58.2&55.2&42.6&34.7&32.3&41.0&54.5&60.9&48.4&44.9\\
LSTKC~\cite{xu2024lstkc}&512&UL&43.6&30.7&45.0&35.0&31.7&24.0&74.4&68.3&66.5&66.2&61.8&59.8&43.4&36.4&29.3&40.3&57.8&66.4&50.4&47.5\\
DJAA (ours)&512&UL&49.0&38.3&\textbf{56.0}&\textbf{44.0}&46.0&36.0&82.9&\textbf{76.7}&69.3&\textbf{69.3}&\textbf{67.3}&\textbf{65.9}&47.5&39.5&31.4&\textbf{48.8}&64.5&\textbf{70.5}&57.1&\textbf{54.3}\\
DJAA* (ours)&512&UL&\textbf{50.9}&\textbf{39.6}&53.4&\textbf{44.0}&\textbf{47.2}&\textbf{37.6}&79.8&71.7&\textbf{70.3}&\textbf{69.7}&66.9&63.8&\textbf{50.1}&\textbf{42.2}&\textbf{36.1}&46.7&\textbf{64.8}&70.5&\textbf{57.7}&54.0\\
\hline
ACL~\cite{zhang2022adaptive}&All&DG&69.2&60.8&60.0&50.0&58.8&48.8&89.9&86.7&78.4&77.8&77.1&76.4&62.4&53.5&65.1&67.3&71.1&78.8&70.2&66.7\\
\hline
\end{tabular}}

\label{table:unseen domains results}
\end{table*}



\subsubsection{Compared methods}
We re-implement 3 types of unsupervised methods, including unsupervised domain adaptation, unsupervised lifelong and supervised domain generalization methods to compare with our method. 

The unsupervised domain adaptive methods include ICE~\cite{Chen_2021_ICE}, CC~\cite{dai2022cluster} and PPRL~\cite{cho2022part}, which are trained sequentially on each seen domain. 
The lifelong methods include three general-purpose lifelong methods (LwF~\cite{Li2018LearningWF}, iCaRL~\cite{Rebuffi2017iCaRLIC} and C$o^2$L~\cite{cha2021co2l}), one backward-compatible class-incremental method CVS~\cite{wan2022continual} and one lifelong ReID method LSTKC~\cite{xu2024lstkc}. LwF is a regularization-based method, which does not store old samples for rehearsal. LwF uses a prediction-level cross-entropy distillation~\cite{hinton2015distilling} between old and new domain models. iCaRL and C$o^2$L are rehearsal-based methods. iCaRL conducts prediction-level distillation on new and stored old images for rehearsal. C$o^2$L proposes an asymmetric supervised contrastive loss and a relation distillation for supervised continual learning. CVS formulates an inter-session data coherence loss and a neighbor-session triplet loss to enhance model coherence during the incremental learning.
For general-purpose methods LwF, iCaRL, C$o^2$L and CVS, we \textbf{combine our unsupervised adaptation module ($\mathcal{L}_{pa}+\mathcal{L}_{ia}$) and the lifelong learning techniques of each paper} to convert these methods to person ReID and conduct a fair comparison with our method. For example, in LwF, we combine our contrastive baseline for learning current domain knowledge and the prediction-level distillation for mitigating the forgetting. To convert the supervised lifelong method LSTKC~\cite{xu2024lstkc} into an unsupervised lifelong method, we combine our unsupervised adaptation module ($\mathcal{L}_{pa}+\mathcal{L}_{ia}$) and the rectification-based knowledge distillation. We also re-implement a supervised domain generalization method ACL~\cite{zhang2022adaptive} to show the upper bound of the generalization ability. The compared lifelong methods, such as LwF~\cite{Li2018LearningWF}, iCaRL~\cite{Rebuffi2017iCaRLIC}, C$o^2$L~\cite{cha2021co2l}, CVS~\cite{wan2022continual} and LSTKC~\cite{xu2024lstkc}, follow the same data augmentation. For ICE~\cite{Chen_2021_ICE}, CC~\cite{dai2022cluster}, PPRL~\cite{cho2022part} and ACL~\cite{zhang2022adaptive}, we directly use their original augmentation setting, including random cropping, flipping and erasing.

\subsection{Seen-domain adaptation performance evaluation}
\begin{figure}[t]
\centering
   \includegraphics[width=1\linewidth]{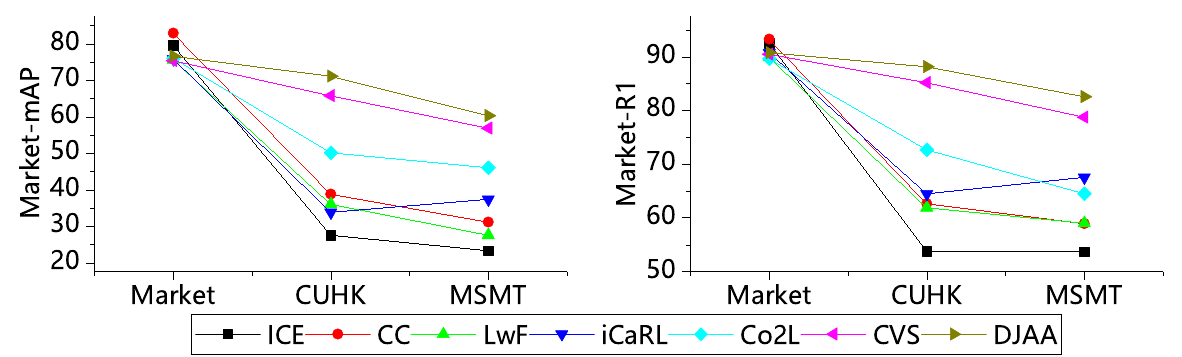}
   \caption{Non-forgetting evaluation with mAP and Rank1 on the first seen domain Market-1501. The training order is Market$\to$Cuhk-Sysu$\to$MSMT17.}
   \label{fig:Non-forgetting evaluation}
\includegraphics[width=1\linewidth]{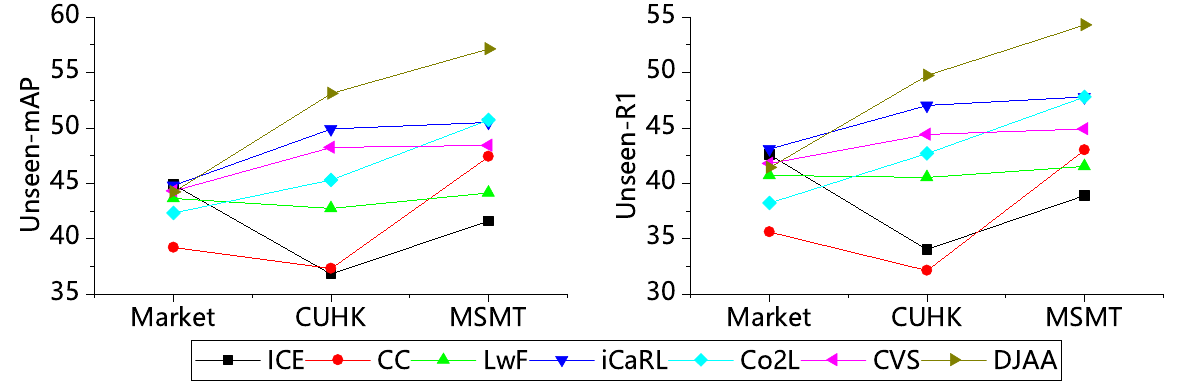}
\caption{Generalization ability evaluation with averaged mAP and averaged Rank1 on all the unseen domains. The training order is Market$\to$Cuhk-Sysu$\to$MSMT17.}
\label{fig:Generalizability evaluation}
\end{figure}

For the one-cycle full set benchmark, we report seen-domain results after the final step in Table~\ref{table:seen domains results}. Designed for maximally learning domain-specific features inside a single domain, regular unsupervised adaptation methods ICE, ClusterContrast (CC) and PPRL cannot learn domain-agnostic generalized features for lifelong ReID. We convert several incremental learning methods, such as LwF, iCaRL, C$o^2$L and CVS, into unsupervised lifelong ReID methods. Among these lifelong methods, the rehearsal-based methods iCaRL and C$o^2$L yield better averaged performance than the pure regularization-based method LwF. We also report the performance of state-of-the-art unsupervised lifelong ReID method CLUDA. Under the unsupervised lifelong setting, our proposed DJAA outperforms the state-of-the-art method CVS by 5.9\% on averaged mAP and 6.7\% on averaged Rank1. We further add camera-aware contrastive loss to reduce the camera style variance. With less camera noise being accumulated into our model, DJAA* achieves higher averaged performance on seen domains. DJAA* does not performs well on CUHK-SYSU, because no camera labels are available on this dataset. We further draw mAP/Rank1 variation curves on the first seen domain Market1501 in Fig.~\ref{fig:Non-forgetting evaluation}, which confirms that our proposed DJAA has a better anti-forgetting capacity. After 3 adaptation steps, our proposed DJAA forgets the least knowledge among all the compared methods. 

For the two-cycle subset benchmark, we compare the adaptation performance after each step in Table~\ref{table:two cycle adaptation performance} and the anti-forgetting performance in Table~\ref{table:two cycle anti-forgetting performance}. As shown in Table~\ref{table:two cycle adaptation performance}, our method DJAA shows superior adaptation performance on the two domain adaptation cycles. These results demonstrate the effectiveness of our proposed adaptation module in adapting a model to new environments. As shown in Table~\ref{table:two cycle anti-forgetting performance}, our method DJAA shows slightly inferior anti-forgetting performance on PersonX, while significantly outperforming CLUDA on Market1501 and CUHK-SYSU. The anti-forgetting performance of DJAA surpasses that of CLUDA on the averaged results over PersonX, Market and Cuhk-Sysu. These results demonstrate the effectiveness of our proposed rehearsal module in alleviating catastrophic forgetting.


\begin{table*}
\centering
\caption{Adaptation performance (\%) evaluation on the two-cycle subset benchmark. The training order is PersonX$\to$Market$\to$Cuhk-Sysu$\to$MSMT17$\to$Market$\to$Cuhk-Sysu$\to$MSMT17.}
\scalebox{0.9}{
\begin{tabular}{l|cc|cc|cc|cc|cc|cc}
\hline
\multirow{2}{*}{Method} & \multicolumn{2}{c|}{Market(t=1)} & \multicolumn{2}{c|}{Cuhk-Sysu(t=2)} & \multicolumn{2}{c|}{MSMT17(t=3)}& \multicolumn{2}{c|}{Market(t=4)} & \multicolumn{2}{c|}{Cuhk-Sysu(t=5)} & \multicolumn{2}{c}{MSMT17(t=6)} \\ \cline{2-13}
\multicolumn{1}{c|}{} & \multicolumn{1}{c}{mAP} & \multicolumn{1}{c|}{R1} &\multicolumn{1}{c}{mAP} & \multicolumn{1}{c|}{R1} &\multicolumn{1}{c}{mAP} & \multicolumn{1}{c|}{R1} &\multicolumn{1}{c}{mAP} & \multicolumn{1}{c|}{R1}&\multicolumn{1}{c}{mAP} & \multicolumn{1}{c|}{R1}&\multicolumn{1}{c}{mAP} & \multicolumn{1}{c}{R1}\\ 
\hline
CLUDA~\cite{huang2022lifelong}&52.96&75.21&71.35&75.38&11.04&28.67&64.87&82.79&78.39&81.86&14.64&33.91\\
DJAA (ours)&\textbf{56.5}&\textbf{79.4}&\textbf{81.1}&\textbf{83.4}&\textbf{16.3}&\textbf{38.5}&\textbf{65.3}&\textbf{84.7}&\textbf{82.9}&\textbf{84.8}&\textbf{16.8}&\textbf{39.5}\\
\hline
\end{tabular}}
\label{table:two cycle adaptation performance}
\end{table*}

\begin{table*}
\centering
\caption{Anti-forgetting performance (\%) evaluation on the two-cycle subset benchmark. The training order is PersonX$\to$Market$\to$Cuhk-Sysu$\to$MSMT17$\to$Market$\to$Cuhk-Sysu$\to$MSMT17.}
\scalebox{0.9}{
\begin{tabular}{l|cc|cc|cc|cc|cc|cc}
\hline
\multirow{3}{*}{Method} & \multicolumn{6}{c|}{t=3} & \multicolumn{6}{c}{t=6}\\ \cline{2-13}
& \multicolumn{2}{c|}{PersonX} & \multicolumn{2}{c|}{Market} & \multicolumn{2}{c|}{Cuhk-Sysu}& \multicolumn{2}{c|}{PersonX} & \multicolumn{2}{c|}{Market} & \multicolumn{2}{c}{Cuhk-Sysu} \\ \cline{2-13}
\multicolumn{1}{c|}{} & \multicolumn{1}{c}{mAP} & \multicolumn{1}{c|}{R1} &\multicolumn{1}{c}{mAP} & \multicolumn{1}{c|}{R1} &\multicolumn{1}{c}{mAP} & \multicolumn{1}{c|}{R1} &\multicolumn{1}{c}{mAP} & \multicolumn{1}{c|}{R1}&\multicolumn{1}{c}{mAP} & \multicolumn{1}{c|}{R1}&\multicolumn{1}{c}{mAP} & \multicolumn{1}{c}{R1}\\ 
\hline
CLUDA~\cite{huang2022lifelong}&\textbf{69.48}&\textbf{80.66}&45.60&69.97&69.44&73.24&\textbf{58.23}&\textbf{76.47}&49.12&76.25&75.43&78.28\\
DJAA (ours)&64.4&80.6&\textbf{48.2}&\textbf{73.6}&\textbf{81.4}&\textbf{84.1}&56.0&75.6&\textbf{55.6}&\textbf{78.2}&\textbf{80.4}&\textbf{82.7}\\
\hline
\end{tabular}}
\label{table:two cycle anti-forgetting performance}
\end{table*}

\begin{table}
\centering
\caption{Generalization ability performance (\%) evaluation on MMP-Retrieval dataset. The training order is PersonX$\to$Market$\to$Cuhk-Sysu$\to$MSMT17$\to$Market$\to$Cuhk-Sysu$\to$MSMT17.}
\scalebox{0.9}{
\begin{tabular}{l|cc}
\hline
\multirow{2}{*}{Method} & \multicolumn{2}{c}{MMP-Retrieval} \\ \cline{2-3}
\multicolumn{1}{c|}{} & \multicolumn{1}{c}{mAP} & \multicolumn{1}{c}{R1} \\ 
\hline
CLUDA~\cite{huang2022lifelong}&41.0& 67.8\\
DJAA (ours)&\textbf{42.7}&\textbf{78.1}\\
\hline
\end{tabular}}
\label{table:DG MMP}
\end{table}

\subsection{Unseen-domain generalization ability evaluation}
For the one-cycle full set benchmark, we report unseen-domain generalization ability results in Table~\ref{table:unseen domains results}. Similar to seen-domain results, ICE, ClusterContrast (CC) and PPRL can hardly learn domain-agnostic generalized features, which leads to low performance on unseen domains. On the contrary, lifelong methods accumulate knowledge from each adapted domain and eventually learn domain-agnostic generalized features. 
With the same baseline, the rehearsal-based methods iCaRL, C$o^2$L and CVS outperform the pure regularization-based method LwF. Under the unsupervised lifelong setting, our proposed DJAA outperforms the second best method C$o^2$L by 6.4\% on averaged mAP and 6.5\% on averaged Rank1. We also add camera-aware contrastive loss to reduce the camera style variance. With camera information, the performance of DJAA* is on par with DJAA on unseen domains, showing that camera information does not bring in any further improvement in generalization ability. 
We further re-implement a domain generalization method ACL, to show the generalization ability of the supervised multi-domain generalization method on unseen domains. ACL is jointly trained on three datasets with human-annotated labels. With more expensive training setup than the unsupervised lifelong learning, ACL shows a strong domain generalization ability. The generalization ability of DG methods is related to the data distribution. If the training datasets and unseen test datasets have significantly different styles, DG methods could fail to generalize to unseen test datasets. Compared to DG, UL methods have better flexibility in data preparation and label annotation. 

For the two-cycle subset benchmark, we compare the generalization performance after the final adaptation step in Table~\ref{table:DG MMP}. Following CLUDA~\cite{huang2022lifelong}, we uniformly downsample the original video sequences of MMPTRACK~\cite{Han_2023_WACV} with a ratio of 128, and divide each downsampled sequence into two halves. On MMP-Retrieval dataset, we report the Rank-1 and mAP scores averaged over all the five scenarios as the final results. Our proposed method DJAA significantly outperforms CLUDA in the Rank-1 score. The results validate the strong domain generalization ability of our method in the repetitive domain adaptation scenario. 

\begin{table*}
\centering
\caption{Backward-compatible performance on Market1501. Full set training order \#1: Market$\to$Cuhk-Sysu$\to$MSMT17. $\mathcal{Q}_1$,$\mathcal{Q}_2$ and $\mathcal{Q}_3$ respectively denote Market1501 query features extracted after step 1, 2 and 3. $\mathcal{G}_1$,$\mathcal{G}_2$ and $\mathcal{G}_3$ respectively denote Market1501 gallery features extracted after step 1, 2 and 3. The colored number is the difference between self-test and cross-test.}
\scalebox{0.9}{
\begin{tabular}{l|cc|cccc|cccc}
\hline
\multirow{3}{*}{Method} & \multicolumn{2}{c|}{Step 1} & \multicolumn{4}{c|}{Step 2} & \multicolumn{4}{c}{Step 3} \\ \cline{2-11}
 & \multicolumn{2}{c|}{Self-Test ($\mathcal{Q}_1,\mathcal{G}_1$)} & \multicolumn{2}{c|}{Self-Test ($\mathcal{Q}_2,\mathcal{G}_2$)} & \multicolumn{2}{c|}{Cross-Test ($\mathcal{Q}_2,\mathcal{G}_1$)} & \multicolumn{2}{c|}{Self-Test ($\mathcal{Q}_3,\mathcal{G}_3$)} & \multicolumn{2}{c}{Cross-Test ($\mathcal{Q}_3,\mathcal{G}_1$)} \\
 & \multicolumn{1}{c}{mAP} & \multicolumn{1}{c|}{R1} & \multicolumn{1}{c}{mAP} & \multicolumn{1}{c|}{R1} & \multicolumn{1}{c}{mAP} & \multicolumn{1}{c|}{R1} & \multicolumn{1}{c}{mAP} & \multicolumn{1}{c|}{R1} & \multicolumn{1}{c}{mAP} & \multicolumn{1}{c}{R1} \\ \hline
ICE~\cite{Chen_2021_ICE}& 80.1&92.8 & 41.4&67.1 & 30.3\textcolor{red}{($\downarrow11.1$)}&49.3\textcolor{red}{($\downarrow17.8$)} & 33.4&65.0 & 3.4\textcolor{red}{($\downarrow30.0$)}&6.3\textcolor{red}{($\downarrow58.7$)}\\
C$o^2$L~\cite{cha2021co2l}&73.8&88.1&47.0&70.2&41.2\textcolor{red}{($\downarrow5.8$)}&58.0\textcolor{red}{($\downarrow12.2$)}&33.7&61.3&9.6\textcolor{red}{($\downarrow24.1$)}&15.9\textcolor{red}{($\downarrow45.4$)}\\
CVS~\cite{wan2022continual}&75.3&90.4&65.7&85.1&61.4\textcolor{red}{($\downarrow4.3$)}&78.3\textcolor{red}{($\downarrow6.8$)}&56.8&78.7&49.6\textcolor{red}{($\downarrow7.2$)}&67.1\textcolor{red}{($\downarrow11.6$)}\\
LSTKC~\cite{xu2024lstkc}&74.2&89.4&63.1&83.9&64.6\textcolor{green}{($\uparrow1.5$)}&83.6\textcolor{red}{($\downarrow0.3$)}&48.8&74.9&38.5\textcolor{red}{($\downarrow10.3$)}&61.3\textcolor{red}{($\downarrow13.6$)}\\
DJAA (ours) & 74.6&90.1 & 70.2&88.3 & \textbf{71.7}\textcolor{green}{($\uparrow1.5$)}&\textbf{89.0}\textcolor{green}{($\uparrow0.7$)} & 59.0&81.0 & \textbf{65.2}\textcolor{green}{($\uparrow6.2$)}&\textbf{85.6}\textcolor{green}{($\uparrow4.6$)}\\
\hline
\end{tabular}}
\label{table:backward compatible order1}
\end{table*}

\subsection{Backward-compatible ability evaluation}
To validate the effectiveness of our proposed method, we compare backward-compatible ability between state-of-the-art methods and our proposed method DJAA in the domain-incremental scenario. As shown in Table~\ref{table:backward compatible order1}, each adaptation step helps to acquire new domain knowledge, while losing previous domain knowledge. ICE is designed for traditional one-step unsupervised domain adaptation, which has the most evident performance drop in the cross-test. C$o^2$L is a class-incremental learning method, which shows better backward-compatible ability than ICE. CVS aims at simultaneously addressing class-incremental learning and backward-compatible learning. Our proposed DJAA uses both image-to-prototype and image-to-image similarity to maximally retain old domain knowledge, which shows optimal backward-compatible ability. The stored gallery features are extracted with the first-domain expert model, which has more discrimination ability than the updated model on the first domain. With compatible gallery features extracted with our DJAA, the cross-test even outperforms the self-test for lifelong person ReID.



\subsection{Ablation study}

\begin{table}
\centering
\caption{Ablation study on the adaptation losses ($\mathcal{L}_{pa}$ and $\mathcal{L}_{ia}$) and the similarity consistency losses ($\mathcal{L}_{ps}$ and $\mathcal{L}_{is}$). We report the averaged results on seen and unseen domains.}
\scalebox{0.9}{
\begin{tabular}{cccc|cc|cc}
\hline
\multirow{2}{*}{$\mathcal{L}_{pa}$}  &\multirow{2}{*}{$\mathcal{L}_{ia}$}  & 
\multirow{2}{*}{$\mathcal{L}_{ps}$}  &\multirow{2}{*}{$\mathcal{L}_{is}$}  & \multicolumn{2}{c|}{Seen} & \multicolumn{2}{c}{Unseen} \\ \cline{5-8}
\multicolumn{1}{c}{} &\multicolumn{1}{c}{} & \multicolumn{1}{c}{} &\multicolumn{1}{c|}{} & \multicolumn{1}{c}{$\bar{s}_{mAP}$} & \multicolumn{1}{c|}{$\bar{s}_{R1}$} & \multicolumn{1}{c}{$\bar{s}_{mAP}$} & \multicolumn{1}{c}{$\bar{s}_{R1}$} \\ \hline
\checkmark&&&&40.2&59.4&41.4&39.7\\
\checkmark&\checkmark&&&41.9&62.7&47.0&44.9\\
\checkmark&\checkmark&\checkmark&&51.9&66.5&52.2&49.5\\
\checkmark&\checkmark&&\checkmark&53.0&70.0&53.3&50.8\\
\checkmark&\checkmark&\checkmark&\checkmark&\textbf{54.9}&\textbf{71.6}&\textbf{57.1}&\textbf{54.3}\\
\hline
\end{tabular}}
\label{table:ablation study}
\end{table}

\begin{figure}
\centering
   \includegraphics[width=1\linewidth]{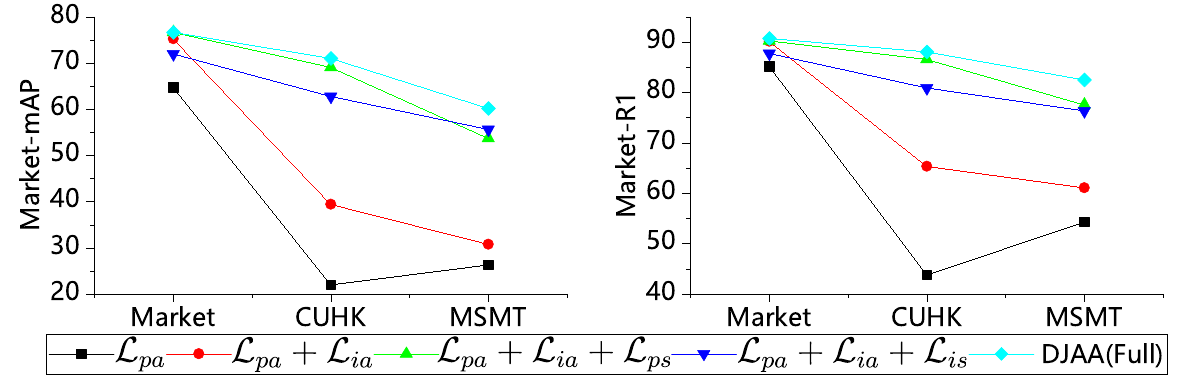}
   \caption{Ablation study on non-forgetting evaluation with mAP and Rank1 on the first seen domain Market-1501.}
\label{fig:Ablation study Non-forgetting evaluation}

 \includegraphics[width=1\linewidth]{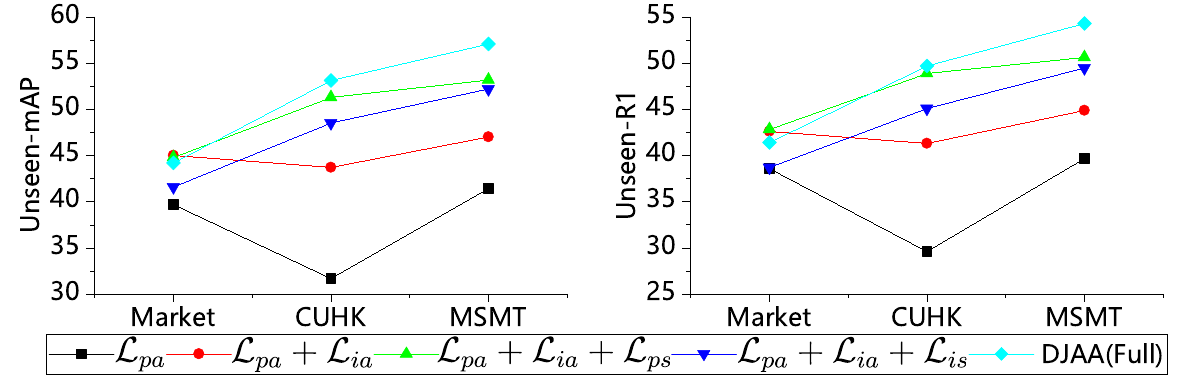}
   \caption{Ablation study on generalization ability evaluation with averaged mAP and averaged Rank1 on all the unseen domains.}
\label{fig:Ablation study Generalizability evaluation}
\end{figure}

\begin{table}
\centering
\caption{Ablation study on the backward compatibility. The Self-Test is reported between $\mathcal{Q}_2$ and $\mathcal{G}_2$. $\mathcal{Q}_2$ is the Market1501 query features extracted after step 2 (Cuhk-Sysu). $\mathcal{G}_2$ is the Market1501 gallery features extracted after step 2. The Cross-Test is reported between $\mathcal{Q}_2$ and $\mathcal{G}_1$. $\mathcal{G}_1$ is the Market1501 query features stored after step 1. }
\scalebox{0.9}{
\begin{tabular}{cccc|cc|cc}
\hline
\multirow{2}{*}{$\mathcal{L}_{pa}$}  &\multirow{2}{*}{$\mathcal{L}_{ia}$}  & 
\multirow{2}{*}{$\mathcal{L}_{ps}$}  &\multirow{2}{*}{$\mathcal{L}_{is}$}  & \multicolumn{2}{c|}{Self-Test($\mathcal{Q}_2,\mathcal{G}_2$)} & \multicolumn{2}{c}{Cross-Test($\mathcal{Q}_2,\mathcal{G}_1$)} \\ \cline{5-8}
\multicolumn{1}{c}{} &\multicolumn{1}{c}{} & \multicolumn{1}{c}{} &\multicolumn{1}{c|}{} &\multicolumn{1}{c}{$mAP$} & \multicolumn{1}{c|}{$R1$} &\multicolumn{1}{c}{$mAP$} & \multicolumn{1}{c}{$R1$}\\ \hline
\checkmark&&&& 19.0& 40.0&9.6\textcolor{red}{($\downarrow9.4$)}&24.1\textcolor{red}{($\downarrow15.9$)}\\
\checkmark&\checkmark&&& 49.5& 72.9&40.1\textcolor{red}{($\downarrow9.4$)}&56.1\textcolor{red}{($\downarrow16.8$)}\\
\checkmark&\checkmark&\checkmark&& 68.2& 86.4&64.5\textcolor{red}{($\downarrow3.7$)}&82.6\textcolor{red}{($\downarrow3.8$)}\\
\checkmark&\checkmark&&\checkmark& 69.0& 87.6&69.9\textcolor{green}{($\uparrow0.9$)}&89.0\textcolor{green}{($\uparrow1.4$)}\\
\checkmark&\checkmark&\checkmark&\checkmark& 70.2& 88.3 &\textbf{71.7}\textcolor{green}{($\uparrow1.5$)}&\textbf{89.0}\textcolor{green}{($\uparrow0.7$)}\\
\hline
\end{tabular}}
\label{table:ablation study backward compatibility}
\end{table}

\subsubsection{Adaptation and rehearsal losses}
To tackle the forgetting problem during the unsupervised domain adaptation, we propose a dual-level joint adaptation and anti-forgetting method. The performance improvement of DJAA over the baseline mainly comes from the combination of the adaptation losses and the rehearsal losses `$\mathcal{L}_{pa}$', `$\mathcal{L}_{ia}$', `$\mathcal{L}_{ps}$' and `$\mathcal{L}_{is}$'.  To validate the effectiveness of each loss, we conduct ablation experiments by gradually adding one of them onto the baseline. In Table~\ref{table:ablation study}, when we only use a prototype adaptation loss `$\mathcal{L}_{pa}$', our model tends to lose most of the previous knowledge. The instance-level adaptation loss `$\mathcal{L}_{ia}$' reduces intra-cluster variance, which prevents the noise accumulation in the multi-step adaptation. The prototype-level similarity consistency loss `$\mathcal{L}_{ps}$' and the instance-level similarity consistency loss `$\mathcal{L}_{is}$' respectively make the image-to-prototype and image-to-image similarity relationships consistent to domain knowledge changes. A cluster prototype contains general information of a cluster, while an instance contains specific information from a single view. As the two consistency losses work on different levels, `$\mathcal{L}_{ps}$' and `$\mathcal{L}_{is}$' are complementary to each other. By combining all the above-mentioned losses, our full DJAA framework yields the highest performance on both seen and unseen domains. We draw the forgetting curve in Fig.~\ref{fig:Ablation study Non-forgetting evaluation} and the generalization ability curve in Fig.~\ref{fig:Ablation study Generalizability evaluation}, which further validate the effectiveness of our proposed losses.

The backward compatible learning aims at maintaining the consistency of representations after training more data, so that previously extracted representations can be directly compared with newly extracted representations. The backward compatibility is strongly correlated with the anti-forgetting ability of a method. We perform an ablation study to validate the effectiveness of our Rehearsal Module in addressing the backward compatibility problem. The difference between the self-test and the cross-test reflects the backward compatibility. As shown in Table~\ref{table:ablation study backward compatibility}, the cross-test performance obviously degrades when we only use the adaptation losses $\mathcal{L}_{pa}$ and $\mathcal{L}_{ia}$. The improvement in the backward compatibility comes mainly from our rehearsal module ($\mathcal{L}_{ps}$ and $\mathcal{L}_{is}$). In particular, the image-to-image similarity consistency loss $\mathcal{L}_{is}$ provides the most significant performance boost.



\begin{table}
\centering
\captionsetup{labelfont={color=red!60!black}, font={color=red!60!black}}
\caption{Ablation study on data augmentation, including random horizontal flipping (Flip), random cropping (Crop), random blurring (Blur) and random erasing (Erase). We report the averaged results on seen and unseen domains.}
\scalebox{0.9}{
\begin{tabular}{ccccc|cc|cc}
\hline
\multirow{2}{*}{Flip}  &\multirow{2}{*}{Crop}  & 
\multirow{2}{*}{Blur}  &\multirow{2}{*}{Erase} &\multirow{2}{*}{ColorJitter} & \multicolumn{2}{c|}{Seen} & \multicolumn{2}{c}{Unseen}  \\ \cline{6-9}
\multicolumn{1}{c}{} &\multicolumn{1}{c}{} &\multicolumn{1}{c}{} & \multicolumn{1}{c}{} &\multicolumn{1}{c|}{} & \multicolumn{1}{c}{$\bar{s}_{mAP}$} & \multicolumn{1}{c|}{$\bar{s}_{R1}$} & \multicolumn{1}{c}{$\bar{s}_{mAP}$} & \multicolumn{1}{c}{$\bar{s}_{R1}$}  \\ \hline
\checkmark&\checkmark&\checkmark&\checkmark&&54.9&\textbf{71.6}&\textbf{57.1}&\textbf{54.3}\\
&\checkmark&\checkmark&\checkmark&&49.8&65.6&51.3&47.8\\
\checkmark&&\checkmark&\checkmark&&52.4&68.5&53.4&49.7\\
\checkmark&\checkmark&&\checkmark&&53.3&69.1&54.3&51.2\\
\checkmark&\checkmark&\checkmark&&&48.1&65.5&49.2&45.5\\
\checkmark&\checkmark&\checkmark&\checkmark&\checkmark&\textbf{55.1}&\textbf{71.6}&\textbf{57.1}&53.2\\
\hline
\end{tabular}}
\label{table:data augmentation ablation study}
\end{table}

\begin{table}
\centering
\caption{Comparison of different data augmentation settings for augmentation consistency regularization.}
\scalebox{0.9}{
\begin{tabular}{ccc|cc|cc}
\hline
\multirow{2}{*}{Loss}  & \multirow{2}{*}{Pred} & \multirow{2}{*}{Ref} & \multicolumn{2}{c|}{Seen} & \multicolumn{2}{c}{Unseen} \\ \cline{4-7}
\multicolumn{1}{c}{} & \multicolumn{1}{c}{} &\multicolumn{1}{c|}{} & \multicolumn{1}{c}{$\bar{s}_{mAP}$} & \multicolumn{1}{c|}{$\bar{s}_{R1}$} & \multicolumn{1}{c}{$\bar{s}_{mAP}$} & \multicolumn{1}{c}{$\bar{s}_{R1}$} \\ \hline
\multirow{3}{*}{$\mathcal{L}_{ps}$}&Weak&Weak&55.2&\textbf{72.3}&56.7&54.1\\
&Strong&Strong&\textbf{55.3}&71.4&56.4&53.8\\
&Strong&Weak&54.9&71.6&\textbf{57.1}&\textbf{54.3}\\\hline
\multirow{3}{*}{$\mathcal{L}_{is}$}&Weak&Weak&48.5&67.9&50.6&48.5\\
&Strong&Strong&54.2&70.7&56.2&53.6\\
&Strong&Weak&\textbf{54.9}&\textbf{71.6}&\textbf{57.1}&\textbf{54.3}\\
\hline
\end{tabular}}

\label{table:Augmentation consistency}
\end{table}

\subsubsection{Data augmentation consistency}
In our method, the data augmentation includes random horizontal flipping, cropping, Gaussian blurring and erasing. These augmentation techniques are chosen to mimic real-world perturbations, such as viewpoint variance, imperfect detection and occlusion. We use these data augmentation techniques to generate augmented views for contrastive learning. By maximizing the similarity between positive views, our model can learn robust representations. As shown in Table~\ref{table:data augmentation ablation study}, all the chosen augmentation techniques contribute to the adaptation performance. Among these augmentation techniques, the random erasing provides the most significant performance improvement, while the random blurring provides the least. Suitable data augmentation should be diverse but realistic, which means augmentation should not bring in distortions absent in the real dataset. The four augmentations (random flipping, cropping, blurring and erasing) are empirically selected. The results in Table~\ref{table:data augmentation ablation study} show that adding color jitter augmentation achieves similar results. Thus, we do not bother adding color jitter.

In addition to regularizing the consistency between two encoders of different steps, we also introduce data augmentation perturbations to further enhance the model robustness. We report the performance of different data augmentation settings in Table~\ref{table:Augmentation consistency}. Different data augmentation settings on the prototype-level similarity consistency loss `$\mathcal{L}_{ps}$' have only a slight influence on the final performance. For the instance-level similarity consistency loss `$\mathcal{L}_{is}$', the influence of the data augmentation setting is more evident. We can observe that data augmentation brings in meaningful perturbations for consistency regularization. Using weakly augmented similarity as reference to regularize the strongly augmented prediction similarity with perturbations is the optimal setting for augmentation consistency regularization.

\subsubsection{Hyperparameter analysis}: We use grid search to set the optimal temperature and weight hyper-parameters to balance our proposed losses. The temperature hyperparameters $\tau_{ps}$ and $\tau_{is}$ control the scale of image-to-prototype and image-to-image similarity. Based on the results in Table~\ref{table: hyperparameters}, we set the temperature hyperparameters $\tau_{ps}=0.1$ and $\tau_{is}=0.2$ in image-to-prototype and image-to-image similarity rehearsal losses, respectively. The weight hyperparameters $\lambda_{ps}$ and $\lambda_{is}$ balance the importance of contrastive adaptation, image-to-prototype and image-to-image similarity losses. Based on the results in Table~\ref{table: hyperparameters}, we set the balancing weight hyperparameters $\lambda_{ps}=10$ and $\lambda_{is}=20$. The exponential moving average hyperparameter $\alpha$ control the the speed of the knowledge accumulation from the online encoder to the momentum encoder. Table~\ref{table: hyperparameters} shows that $\alpha=0.999$ is the optimal setting for our framework. The hyperparameters are tuned in the training order Market$\to$Cuhk-Sysu$\to$MSMT17. The tuned hyperparameters are kept the same for the second training order MSMT17$\to$Market$\to$Cuhk-Sysu, which validates the effectiveness of these hyperparameters.

\begin{table}
\centering
\caption{Comparison of different values for training hyperparameters. $\tau_{ps}$ and $\tau_{is}$ are temperature hyperparameters. $\lambda_{ps}$ and $\lambda_{is}$ are weight balancing hyperparameters. $\alpha$ is the exponential moving average (EMA) hyperparameter.}
\scalebox{0.9}{
\setlength{\tabcolsep}{1pt}
\begin{tabular}{c|cc|cc}
\hline
\multirow{2}{*}{$\tau_{ps}$}  & \multicolumn{2}{c}{Seen} & \multicolumn{2}{|c}{Unseen} \\ \cline{2-5}
\multicolumn{1}{c|}{} & \multicolumn{1}{c}{$\bar{s}_{mAP}$} & \multicolumn{1}{c|}{$\bar{s}_{Rank1}$} & \multicolumn{1}{c}{$\bar{s}_{mAP}$} & \multicolumn{1}{c}{$\bar{s}_{Rank1}$} \\ \hline
0.05&52.5&68.5&53.9&51.3\\
0.1&54.9&\textbf{71.6}&\textbf{57.1}&\textbf{54.3}\\
0.2&\textbf{55.4}&\textbf{71.6}&56.1&52.8\\
\hline
\end{tabular}}
\scalebox{0.9}{
\setlength{\tabcolsep}{1pt}
\begin{tabular}{c|cc|cc}
\hline
\multirow{2}{*}{$\tau_{is}$}  & \multicolumn{2}{c}{Seen} & \multicolumn{2}{|c}{Unseen} \\ \cline{2-5}
\multicolumn{1}{c|}{} & \multicolumn{1}{c}{$\bar{s}_{mAP}$} & \multicolumn{1}{c|}{$\bar{s}_{Rank1}$} & \multicolumn{1}{c}{$\bar{s}_{mAP}$} & \multicolumn{1}{c}{$\bar{s}_{Rank1}$} \\ \hline
0.1&53.5&69.6&52.5&49.5\\
0.2&\textbf{54.9}&\textbf{71.6}&\textbf{57.1}&\textbf{54.3}\\
0.3&47.5&65.6&48.3&44.3\\
\hline
\end{tabular}}
\vspace{1em}
\scalebox{0.9}{
\setlength{\tabcolsep}{1pt}
\begin{tabular}{c|cc|cc}
\hline
\multirow{2}{*}{$\lambda_{ps}$}  & \multicolumn{2}{c}{Seen} & \multicolumn{2}{|c}{Unseen} \\ \cline{2-5}
\multicolumn{1}{c|}{} & \multicolumn{1}{c}{$\bar{s}_{mAP}$} & \multicolumn{1}{c|}{$\bar{s}_{Rank1}$} & \multicolumn{1}{c}{$\bar{s}_{mAP}$} & \multicolumn{1}{c}{$\bar{s}_{Rank1}$} \\ \hline
5&54.6&70.9&53.9&50.3\\
10&\textbf{54.9}&\textbf{71.6}&\textbf{57.1}&\textbf{54.3}\\
15&47.9&64.4&49.0&45.4\\
\hline
\end{tabular}}
\scalebox{0.9}{
\setlength{\tabcolsep}{1pt}
\begin{tabular}{c|cc|cc}
\hline
\multirow{2}{*}{$\lambda_{is}$}  & \multicolumn{2}{c}{Seen} & \multicolumn{2}{|c}{Unseen} \\ \cline{2-5}
\multicolumn{1}{c|}{} & \multicolumn{1}{c}{$\bar{s}_{mAP}$} & \multicolumn{1}{c|}{$\bar{s}_{Rank1}$} & \multicolumn{1}{c}{$\bar{s}_{mAP}$} & \multicolumn{1}{c}{$\bar{s}_{Rank1}$} \\ \hline
10&54.2&69.6&54.0&50.7\\
20&\textbf{54.9}&\textbf{71.6}&\textbf{57.1}&\textbf{54.3}\\
30&52.9&71.1&55.2&52.4\\
\hline
\end{tabular}}
\vspace{1em}
\scalebox{0.9}{
\begin{tabular}{c|cc|cc}
\hline
\multirow{2}{*}{$\alpha$}  & \multicolumn{2}{c}{Seen} & \multicolumn{2}{|c}{Unseen} \\ \cline{2-5}
\multicolumn{1}{c|}{} & \multicolumn{1}{c}{$\bar{s}_{mAP}$} & \multicolumn{1}{c|}{$\bar{s}_{Rank1}$} & \multicolumn{1}{c}{$\bar{s}_{mAP}$} & \multicolumn{1}{c}{$\bar{s}_{Rank1}$} \\ \hline
    0.99&53.4&70.1&53.2&50.8\\
    0.999&\textbf{54.9}&\textbf{71.6}&\textbf{57.1}&\textbf{54.3}\\
    0.9999&38.4&51.2&42.4&40.3\\
\hline
\end{tabular}}
\label{table: hyperparameters}
\end{table}

\label{sec:Parameter analysis}

\subsection{Discussion}

\begin{table*}
\centering
\caption{Seen-domain results (\%) of unsupervised (U) and unsupervised lifelong (UL) methods on one-cycle full set benchmark under the training order MSMT17$\to$Market$\to$Cuhk-Sysu. * refers to using the camera loss~Eq.(\ref{equa:cameraloss}) to reduce intra-cluster variance. The best performance is marked in bold.}
\scalebox{0.9}{
\begin{tabular}{l|c|c|cc|cc|cc|cc}
\hline
\multirow{2}{*}{Method} & Memory & \multirow{2}{*}{Type} & \multicolumn{2}{c|}{MSMT17} & \multicolumn{2}{c|}{Market} & \multicolumn{2}{c|}{Cuhk-Sysu} & \multicolumn{2}{c}{Average} \\ \cline{4-11}
\multicolumn{1}{c|}{} &\multicolumn{1}{c|}{size}&\multicolumn{1}{c|}{}& \multicolumn{1}{c}{mAP} & \multicolumn{1}{c|}{Rank1} &\multicolumn{1}{c}{mAP} & \multicolumn{1}{c|}{Rank1} &\multicolumn{1}{c}{mAP} & \multicolumn{1}{c|}{Rank1} &\multicolumn{1}{c}{$\bar{s}_{mAP}$} & \multicolumn{1}{c}{$\bar{s}_{Rank1}$}  \\ 
\hline
ICE~\cite{Chen_2021_ICE}&0&U&5.4&14.0&42.1&63.4&\textbf{82.5}&\textbf{84.2}&43.3&53.9\\
CC~\cite{dai2022cluster}&0&U&4.8&14.0&40.9&64.6&74.1&76.9&39.9&51.8\\
PPRL~\cite{cho2022part}&0&U&4.4&12.9&45.4&70.5&80.2&83.0&43.3&55.5\\
LwF~\cite{Li2018LearningWF}&0&UL&6.4&20.8&37.8&66.0&74.2&77.3&39.5&54.7\\
iCaRL~\cite{Rebuffi2017iCaRLIC}&512&UL&6.5&20.3&40.0&68.7&75.7&78.7&40.7&55.9\\
C$o^2$L~\cite{cha2021co2l}&512&UL&7.9&24.4&42.8&69.5&79.8&82.3&43.5&58.8\\
CLUDA~\cite{huang2022lifelong}&512&UL&13.9&31.5&50.1&77.3&82.1&84.2&48.7&64.3\\
CVS~\cite{wan2022continual}&512&UL&14.6&39.6&53.3&79.1&75.2&77.6&47.7&65.4\\
LSTKC~\cite{xu2024lstkc}&512&UL&11.7&32.1&47.8&75.4&77.6&80.4&45.7&63.6\\
DJAA (ours)&512&UL&18.1&43.8&53.7&79.0&82.0&84.0&51.3&68.9\\
DJAA* (ours)&512&UL&\textbf{20.2}&\textbf{48.5}&\textbf{59.3}&\textbf{84.1}&79.2&81.9&\textbf{52.9}&\textbf{71.5}\\
\hline
\end{tabular}}
\label{table:order 2 seen}
\end{table*}

\begin{table*}
\centering
\caption{Unseen-domain results (\%) of unsupervised (U), unsupervised lifelong (UL) and domain generalization (DG) methods on one-cycle full set benchmark. The training order is MSMT17$\to$Market$\to$Cuhk-Sysu. * refers to using the camera loss~Eq.(\ref{equa:cameraloss}) to reduce intra-cluster variance. The best unsupervised performance is marked in bold.}
\scalebox{0.9}{
\setlength{\tabcolsep}{3pt}
\begin{tabular}{l|c|c|cc|cc|cc|cc|cc|cc|cc|cc|cc|cc}
\hline
\multirow{2}{*}{Method} & Memory&\multirow{2}{*}{Type}& \multicolumn{2}{c|}{VIPeR} & \multicolumn{2}{c|}{PRID} & \multicolumn{2}{c|}{GRID} & \multicolumn{2}{c|}{iLIDS} & \multicolumn{2}{c|}{CUHK01}& \multicolumn{2}{c|}{CUHK02}& \multicolumn{2}{c|}{SenseReID}& \multicolumn{2}{c|}{CUHK03}& \multicolumn{2}{c|}{3DPeS}& \multicolumn{2}{c}{Average} \\ \cline{4-23}
\multicolumn{1}{c|}{}&\multicolumn{1}{c|}{Size}&\multicolumn{1}{c|}{}& \multicolumn{1}{c}{mAP} & \multicolumn{1}{c|}{R1} &\multicolumn{1}{c}{mAP} & \multicolumn{1}{c|}{R1} &\multicolumn{1}{c}{mAP} & \multicolumn{1}{c|}{R1} &\multicolumn{1}{c}{mAP} & \multicolumn{1}{c|}{R1} &\multicolumn{1}{c}{mAP} & \multicolumn{1}{c|}{R1} &\multicolumn{1}{c}{mAP} & \multicolumn{1}{c|}{R1} &\multicolumn{1}{c}{mAP} & \multicolumn{1}{c|}{R1} &\multicolumn{1}{c}{mAP} & \multicolumn{1}{c|}{R1} &\multicolumn{1}{c}{mAP} & \multicolumn{1}{c|}{R1} &\multicolumn{1}{c}{$\bar{s}_{mAP}$} & \multicolumn{1}{c}{$\bar{s}_{R1}$}  \\ 
\hline
ICE~\cite{Chen_2021_ICE}&0&U&41.3&31.6&32.8&22.0&31.1&20.0&72.1&61.7&47.6&47.1&51.3&47.3&36.7&29.1&13.6&32.7&47.1&55.5&41.5&38.6\\
CC~\cite{dai2022cluster}&0&U&36.7&25.9&35.0&26.0&29.2&20.8&64.1&55.0&40.7&38.5&45.1&42.1&31.7&24.4&18.0&17.1&46.4&56.8&38.5&34.1\\
PPRL~\cite{cho2022part}&0&U&38.7&30.4&25.4&16.0&28.3&20.8&69.6&60.0&47.4&47.5&47.5&43.1&29.6&23.5&14.4&13.5&41.8&53.9&38.1&34.3\\
LwF~\cite{Li2018LearningWF}&0&UL&40.5&30.1&36.6&28.0&36.6&29.6&73.0&61.7&45.9&45.5&51.0&47.3&33.7&26.7&19.2&30.8&50.6&58.6&43.0&39.8\\
iCaRL~\cite{Rebuffi2017iCaRLIC}&512&UL&40.3&29.7&36.0&25.0&36.0&26.4&74.4&66.7&48.2&46.6&49.6&45.8&35.0&28.0&18.6&32.2&52.6&60.5&43.4&40.1\\
C$o^2$L~\cite{cha2021co2l}&512&UL&41.8&31.6&50.2&38.0&38.4&27.2&79.0&70.0&56.6&55.6&57.0&55.0&38.2&31.8&20.9&36.7&56.4&65.5&48.7&45.7\\
CVS~\cite{wan2022continual}&512&UL&41.0&32.3&43.5&33.0&\textbf{41.3}&31.2&\textbf{85.0}&\textbf{80.0}&61.0&57.8&59.1&56.3&40.4&32.7&29.4&36.3&56.4&64.1&50.8&47.1\\
LSTKC~\cite{xu2024lstkc}&512&UL&39.1&27.8&45.3&35.0&37.8&28.8&77.4&68.3&61.6&61.7&60.1&59.6&38.2&29.9&23.0&36.5&51.6&61.4&48.2&45.5\\
DJAA (ours)&512&UL&\textbf{47.6}&\textbf{36.4}&\textbf{51.8}&\textbf{41.0}&40.6&\textbf{32.8}&82.5&75.0&69.1&68.9&\textbf{67.4}&\textbf{65.7}&44.9&\textbf{37.6}&25.5&\textbf{45.2}&57.2&64.1&54.1&\textbf{51.9}\\
DJAA* (ours)&512&UL&46.1&33.9&48.5&37.0&40.5&30.4&80.5&71.7&\textbf{70.7}&\textbf{70.2}&\textbf{67.4}&65.3&\textbf{45.0}&37.1&\textbf{31.0}&43.6&\textbf{59.3}&\textbf{66.4}&\textbf{54.3}&50.6\\
\hline
ACL~\cite{zhang2022adaptive}&All&DG&69.2&60.8&60.0&50.0&58.8&48.8&89.9&86.7&78.4&77.8&77.1&76.4&62.4&53.5&65.1&67.3&71.1&78.8&70.2&66.7\\
\hline
\end{tabular}}

\label{table:order 2 unseen}
\end{table*}

\begin{table*}
\centering
\caption{Backward-compatible performance on MSMT17. Full set training order \#2: MSMT17$\to$Market$\to$Cuhk-Sysu. $\mathcal{Q}_1$,$\mathcal{Q}_2$ and $\mathcal{Q}_3$ respectively denote MSMT17 query features extracted after step 1, 2 and 3. $\mathcal{G}_1$,$\mathcal{G}_2$ and $\mathcal{G}_3$ respectively denote MSMT17 gallery features extracted after step 1, 2 and 3. The colored number is the difference between self-test and cross-test.}
\scalebox{0.9}{
\begin{tabular}{l|cc|cccc|cccc}
\hline
\multirow{3}{*}{Method} & \multicolumn{2}{c|}{Step 1} & \multicolumn{4}{c|}{Step 2} & \multicolumn{4}{c}{Step 3} \\ \cline{2-11}
 & \multicolumn{2}{c|}{Self-Test ($\mathcal{Q}_1,\mathcal{G}_1$)} & \multicolumn{2}{c|}{Self-Test ($\mathcal{Q}_2,\mathcal{G}_2$)} & \multicolumn{2}{c|}{Cross-Test ($\mathcal{Q}_2,\mathcal{G}_1$)} & \multicolumn{2}{c|}{Self-Test ($\mathcal{Q}_3,\mathcal{G}_3$)} & \multicolumn{2}{c}{Cross-Test ($\mathcal{Q}_3,\mathcal{G}_1$)} \\
 & \multicolumn{1}{c}{mAP} & \multicolumn{1}{c|}{R1} & \multicolumn{1}{c}{mAP} & \multicolumn{1}{c|}{R1} & \multicolumn{1}{c}{mAP} & \multicolumn{1}{c|}{R1} & \multicolumn{1}{c}{mAP} & \multicolumn{1}{c|}{R1} & \multicolumn{1}{c}{mAP} & \multicolumn{1}{c}{R1} \\ \hline
ICE~\cite{Chen_2021_ICE}&32.7&65.7&6.6&20.1&3.8\textcolor{red}{($\downarrow2.8$)}&9.3\textcolor{red}{($\downarrow10.8$)}&7.6&22.6&1.0\textcolor{red}{($\downarrow6.6$)}&2.7\textcolor{red}{($\downarrow19.9$)}\\
C$o^2$L~\cite{cha2021co2l}&25.8&57.6&7.0&20.9&2.2\textcolor{red}{($\downarrow4.8$)}&6.1\textcolor{red}{($\downarrow14.8$)}&7.7&23.8&0.2\textcolor{red}{($\downarrow7.5$)}&0.3\textcolor{red}{($\downarrow23.5$)}\\
CVS~\cite{wan2022continual}&27.3&59.2&18.3&44.7&19.5\textcolor{green}{($\uparrow1.2$)}&45.4\textcolor{green}{($\uparrow0.7$)}&14.6&39.6&15.3\textcolor{green}{($\uparrow0.7$)}&36.6\textcolor{red}{($\downarrow3.0$)}\\
LSTKC~\cite{xu2024lstkc}&28.0&59.8&17.1&41.2&19.7\textcolor{green}{($\uparrow2.6$)}&44.7\textcolor{green}{($\uparrow3.5$)}&11.7&32.1&12.2\textcolor{green}{($\uparrow0.5$)}&30.4\textcolor{red}{($\downarrow1.7$)}\\
DJAA (ours)&27.1&59.3&21.6&49.4&\textbf{24.4}\textcolor{green}{($\uparrow2.8$)}&\textbf{53.9}\textcolor{green}{($\uparrow4.5$)}&18.1&43.8&\textbf{20.6}\textcolor{green}{($\uparrow2.5$)}&\textbf{48.3}\textcolor{green}{($\uparrow4.5$)}\\
\hline
\end{tabular}}
\label{table:backward compatible order2}
\end{table*}

\begin{figure}
\centering
\includegraphics[width=1\linewidth]{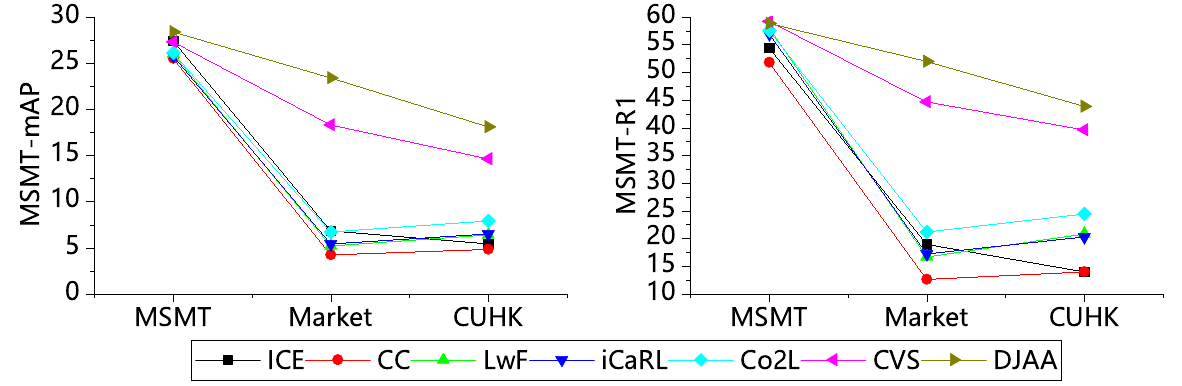}
\caption{Non-forgetting evaluation with mAP and Rank1 on the first seen domain MSMT17. The training order is MSMT17$\to$Market501$\to$Cuhk-Sysu.}
\label{fig:Non-forgetting evaluation order2}
\includegraphics[width=1\linewidth]{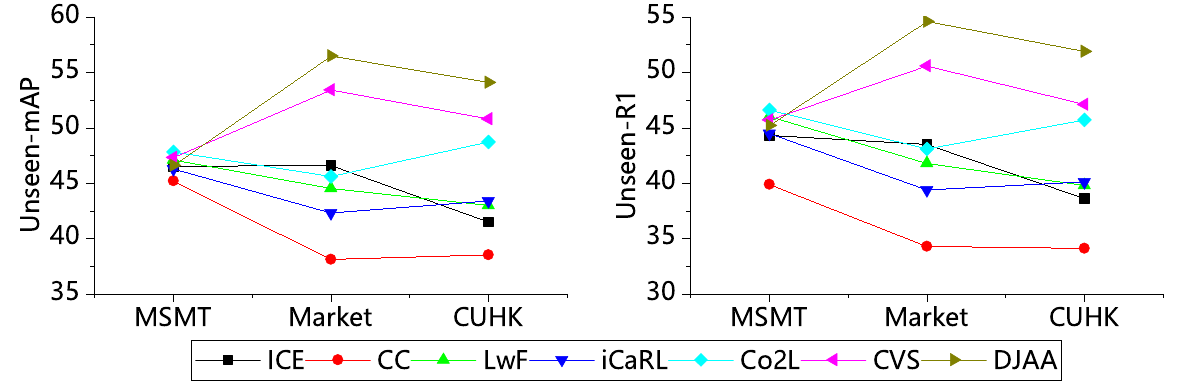}
\caption{Generalization ability evaluation with averaged mAP and averaged Rank1 on all the unseen domains. The training order is MSMT17$\to$Market501$\to$Cuhk-Sysu.}
\label{fig:Generalizability evaluation order2}
\end{figure}

\subsubsection{Training order} As datasets are of different scales and diversity, it is meaningful to know whether the training order has a significant influence on the final results. Our primary training order starts from a medium domain Market and ends with the largest domain MSMT17. However, it is hard to control the order of upcoming domains in the real world. We test a second order MSMT17$\to$Market$\to$Cuhk-Sysu that starts from the largest domain MSMT17. In the second training order, it is easier to forget more knowledge on the largest domain MSMT17. Consequently, the performance under the training order \#2 is slightly inferior to that of the training order \#1. As shown in Table~\ref{table:order 2 seen}, DJAA still significantly outperforms the state-of-the-art methods CLUDA and CVS on seen domains. In the meantime, as shown in Table~\ref{table:order 2 unseen}, DJAA also outperforms state-of-the-art methods on unseen domains. We provide a forgetting curve on the seen domains in Fig.~\ref{fig:Non-forgetting evaluation order2} and a generalization ability on unseen domains in Fig.~\ref{fig:Generalizability evaluation order2}. Compared to previous methods, our proposed DJAA shows both strong anti-forgetting and generalization abilities. We further compare backward-compatible ability between state-of-the-art methods and our proposed method DJAA in the second training order. As shown in Table~\ref{table:order 2 seen}, DJAA consistently shows superior backward-compatible ability over previous methods in the second training order.

\begin{table}
\centering
\caption{Comparison of memory buffer updating strategies. `Random' refers to randomly sample $n_{new}$ images from old and current domains. `ID-wise' refers to randomly sample $n_{new}$ clusters and one random image per cluster. `ours' refers to sample $n_{new}$ clusters with the largest cluster sample numbers and one image nearest to the prototype.}
\scalebox{0.9}{
\begin{tabular}{c|cc|cc}
\hline
\multirow{2}{*}{Method}  & \multicolumn{2}{c}{Seen} & \multicolumn{2}{|c}{Unseen} \\ \cline{2-5}
\multicolumn{1}{c|}{} & \multicolumn{1}{c}{$\bar{s}_{mAP}$} & \multicolumn{1}{c|}{$\bar{s}_{Rank1}$} & \multicolumn{1}{c}{$\bar{s}_{mAP}$} & \multicolumn{1}{c}{$\bar{s}_{Rank1}$} \\ \hline
    Random&53.8&69.8&53.9&50.6\\
    ID-wise&54.6&71.1&55.8&53.1\\
    ours&\textbf{54.9}&\textbf{71.6}&\textbf{57.1}&\textbf{54.3}\\
    \hline
    \end{tabular}}
    \label{table:sample selection}
\end{table}

\begin{table}
\centering
\caption{Number of images per pseudo identity in the memory. `Baseline' refers to only use the adaptation module, \ie, $\mathcal{L}_{pa}+\mathcal{L}_{ia}$, which does not store any sample for rehearsal. }
\scalebox{1}{
\begin{tabular}{c|cc|cc}
\hline
\multirow{1}{*}{Buffer Size}  & \multicolumn{2}{c}{Seen} & \multicolumn{2}{|c}{Unseen} \\ \cline{2-5}
\multicolumn{1}{c|}{$n_{mem}$} & \multicolumn{1}{c}{$\bar{s}_{mAP}$} & \multicolumn{1}{c|}{$\bar{s}_{Rank1}$} & \multicolumn{1}{c}{$\bar{s}_{mAP}$} & \multicolumn{1}{c}{$\bar{s}_{Rank1}$} \\ \hline
    Baseline&41.9&62.7&47.0&44.9\\\hline
    32&35.8&52.9&38.8&35.3\\
    64&40.2&56.4&41.5&38.8\\
    128&49.8&67.3&51.8&49.3\\
    256&52.5&69.6&54.2&51.5\\
    512&\textbf{54.9}&\textbf{71.6}&\textbf{57.1}&\textbf{54.3}\\
\hline
\end{tabular}}

\label{table:n mem}
\end{table}

\subsubsection{Data selection for memory buffer update}
\label{Data selection for memory buffer update}
To update the image memory at the end of each step, our strategy is to first select $n_{new}$ clusters with largest cluster sample numbers. The prototypes of selected clusters are used to update the memory buffer. Then, we select one image that is closest to its cluster prototype. In this way, we select the representative clusters for our image-to-prototype similarity consistency loss. As shown in Table~\ref{table:sample selection}, we compare three possible memory buffer updating strategies. `Random' refers to randomly sample $n_{new}$ images from the current domain, which neglects the clustering information. `ID-wise' refers to randomly sample $n_{new}$ clusters and one image per cluster, which has better data diversity than `Random'. `ours' refers to our proposed strategy in Section~\ref{sec:Memory Buffer Update} that selects representative clusters and credible images, which brings in the best performance on both seen and unseen domains. 

\subsubsection{Memory buffer size}
We build a hybrid memory buffer to store cluster prototypes and image samples for the rehearsal module of our method DJAA. With a default memory size $n_{mem}=512$, our image memory stores approximately $512$ images $\approx 0.8\%$ of all the training images (Market, Cuhk-Sysu and MSMT17). Meanwhile, our prototype memory stores approximately $512$ prototype vectors (dimension $1\times2048\times1\times1$)$\approx$4.2 MB, which is negligible compared with storing dataset images (for example, MSMT17$\approx$2.5 GB). 

To further evaluate the dependency on the memory size, we vary the value of $n_{mem}$ and report the results in Table~\ref{table:n mem}. If the buffer size is limited to an extremely small number, such as 32 or 64, the results are even lower than our adaptation module baseline. When the buffer size is greater than or equals 128, the results are higher than the baseline. We can observe that the more data we store, the higher performance DJAA can achieve. As shown in Table~\ref{table:seen domains results}, CLUDA~\cite{huang2022lifelong} achieves 46.4\% mAP and 62.8\% mAP with a buffer size of 512. It is worth mentioning that our method achieves higher performance (49.8\% mAP and 67.3\% mAP) than CLUDA~\cite{huang2022lifelong}, when the buffer size equals 128.

\subsubsection{Efficiency of backward compatibility}
Backward compatible learning aims to maintain the consistency of representations after training more data, so that we do not need to re-extract gallery features after each adaptation step. As shown in Table~\ref{table:backward compatible order1}, with Market gallery features extracted at step 1, ICE has a cross-test performance 30.3\% mAP / 49.3\% Rank1 at step 2 and 3.4\% mAP / 6.3\% Rank1 at step 3. In such case, Market gallery features need to be extracted at steps 1, 2 and 3, and Cuhk-Sysu gallery features needs to be extracted at steps 2 and 3. In contrast, reinforced by the feature back-compatibility, Market and Cuhk-Sysu gallery features need to be extracted only once in our method, at the 1$^{st}$ and 2$^{nd}$ step respectively. For example, with stored Market gallery features, DJAA has a cross-test performance 71.7\% mAP / 89.0\% Rank1 at step 2 and 65.2\% mAP / 85.6\% Rank1 at step 3. A real-world lifelong ReID system may involve more adaptation steps, where a backward compatible method can avoid repetitive gallery feature extraction and fasten the inference.

\section{Conclusion}
In this paper, we propose an anti-forgetting adaptation method for unsupervised person ReID. The traditional unsupervised domain adaptation methods aim at obtaining the optimal performance on a fixed target domain, which shows poor anti-forgetting, generalization and backward-compatible ability. To tackle the three problems at once, we propose a Dual-level Joint Adaptation and Anti-forgetting (DJAA) method, which mainly consists of an adaptation module and a rehearsal module. In the adaptation module, we leverage a prototype-level contrastive loss and an instance-level contrastive loss to maximize the positive view similarity for learning new domain features. In the rehearsal module, our method regularizes the image-to-prototype and image-to-image similarity across domains to mitigate the knowledge forgetting. In comparison with previous lifelong methods, our proposed DJAA significantly improves the non-forgetting ability on seen domains and better generalization ability on unseen domains.


%




\ifCLASSOPTIONcaptionsoff
  \newpage
\fi



%
\bibliographystyle{IEEEtran}
\bibliography{egbib}

\begin{thebibliography}{10}
\providecommand{\url}[1]{#1}
\csname url@samestyle\endcsname
\providecommand{\newblock}{\relax}
\providecommand{\bibinfo}[2]{#2}
\providecommand{\BIBentrySTDinterwordspacing}{\spaceskip=0pt\relax}
\providecommand{\BIBentryALTinterwordstretchfactor}{4}
\providecommand{\BIBentryALTinterwordspacing}{\spaceskip=\fontdimen2\font plus
\BIBentryALTinterwordstretchfactor\fontdimen3\font minus \fontdimen4\font\relax}
\providecommand{\BIBforeignlanguage}[2]{{%
\expandafter\ifx\csname l@#1\endcsname\relax
\typeout{** WARNING: IEEEtran.bst: No hyphenation pattern has been}%
\typeout{** loaded for the language `#1'. Using the pattern for}%
\typeout{** the default language instead.}%
\else
\language=\csname l@#1\endcsname
\fi
#2}}
\providecommand{\BIBdecl}{\relax}
\BIBdecl

\bibitem{vandermaaten08a}
L.~van~der Maaten and G.~Hinton, ``Visualizing data using t-sne,'' \emph{JMLR}, 2008.

\bibitem{Chen_2021_ICE}
H.~Chen, B.~Lagadec, and F.~Bremond, ``Ice: Inter-instance contrastive encoding for unsupervised person re-identification,'' in \emph{ICCV}, 2021.

\bibitem{Ye_2021_reidsurvey}
M.~Ye, J.~Shen, G.~Lin, T.~Xiang, L.~Shao, and S.~C.~H. Hoi, ``Deep learning for person re-identification: A survey and outlook,'' \emph{IEEE TPAMI}, 2021.

\bibitem{zheng2019joint}
Z.~Zheng, X.~Yang, Z.~Yu, L.~Zheng, Y.~Yang, and J.~Kautz, ``Joint discriminative and generative learning for person re-identification,'' in \emph{CVPR}, 2019.

\bibitem{He_2021_ICCV}
S.~He, H.~Luo, P.~Wang, F.~Wang, H.~Li, and W.~Jiang, ``Transreid: Transformer-based object re-identification,'' in \emph{ICCV}, 2021.

\bibitem{zhong2020learning}
Z.~Zhong, L.~Zheng, Z.~Luo, S.~Li, and Y.~Yang, ``Learning to adapt invariance in memory for person re-identification,'' \emph{IEEE TPAMI}, 2020.

\bibitem{pu_cvpr2021}
N.~Pu, W.~Chen, Y.~Liu, E.~M. Bakker, and M.~S. Lew, ``Lifelong person re-identification via adaptive knowledge accumulation,'' in \emph{CVPR}, 2021.

\bibitem{Wu2021GeneralisingWF}
G.~Wu and S.~Gong, ``Generalising without forgetting for lifelong person re-identification,'' in \emph{AAAI}, 2021.

\bibitem{van2022three}
G.~M. van~de Ven, T.~Tuytelaars, and A.~S. Tolias, ``Three types of incremental learning,'' \emph{Nature Machine Intelligence}, pp. 1--13, 2022.

\bibitem{Song2019GeneralizablePR}
J.~Song, Y.~Yang, Y.-Z. Song, T.~Xiang, and T.~M. Hospedales, ``Generalizable person re-identification by domain-invariant mapping network,'' \emph{CVPR}, 2019.

\bibitem{shen2020towards}
Y.~Shen, Y.~Xiong, W.~Xia, and S.~Soatto, ``Towards backward-compatible representation learning,'' in \emph{CVPR}, 2020.

\bibitem{wan2022continual}
T.~S. Wan, J.-C. Chen, T.-Y. Wu, and C.-S. Chen, ``Continual learning for visual search with backward consistent feature embedding,'' in \emph{CVPR}, 2022.

\bibitem{tang2021gradient}
S.~Tang, P.~Su, D.~Chen, and W.~Ouyang, ``Gradient regularized contrastive learning for continual domain adaptation,'' in \emph{AAAI}, 2021.

\bibitem{rostami2021lifelong}
M.~Rostami, ``Lifelong domain adaptation via consolidated internal distribution,'' \emph{NeurIPS}, 2021.

\bibitem{lee2022negative}
H.~Lee, S.~Eum, and H.~Kwon, ``Negative samples are at large: Leveraging hard-distance elastic loss for re-identification,'' in \emph{ECCV}, 2022.

\bibitem{zhou2023adaptive}
X.~Zhou, Y.~Zhong, Z.~Cheng, F.~Liang, and L.~Ma, ``Adaptive sparse pairwise loss for object re-identification,'' in \emph{CVPR}, 2023.

\bibitem{zhang2023pha}
G.~Zhang, Y.~Zhang, T.~Zhang, B.~Li, and S.~Pu, ``Pha: Patch-wise high-frequency augmentation for transformer-based person re-identification,'' in \emph{CVPR}, 2023.

\bibitem{liu2019adaptive}
J.~Liu, Z.-J. Zha, D.~Chen, R.~Hong, and M.~Wang, ``Adaptive transfer network for cross-domain person re-identification,'' in \emph{CVPR}, 2019.

\bibitem{zheng2021exploiting}
K.~Zheng, C.~Lan, W.~Zeng, Z.~Zhang, and Z.-J. Zha, ``Exploiting sample uncertainty for domain adaptive person re-identification,'' in \emph{AAAI}, 2021.

\bibitem{zheng2021group}
K.~Zheng, W.~Liu, L.~He, T.~Mei, J.~Luo, and Z.-J. Zha, ``Group-aware label transfer for domain adaptive person re-identification,'' in \emph{CVPR}, 2021.

\bibitem{ge2020self}
Y.~Ge, F.~Zhu, D.~Chen, R.~Zhao, and H.~Li, ``Self-paced contrastive learning with hybrid memory for domain adaptive object re-id,'' in \emph{NeurIPS}, 2020.

\bibitem{cho2022part}
Y.~Cho, W.~J. Kim, S.~Hong, and S.-E. Yoon, ``Part-based pseudo label refinement for unsupervised person re-identification,'' in \emph{CVPR}, 2022.

\bibitem{Jin_2020_CVPR}
X.~Jin, C.~Lan, W.~Zeng, Z.~Chen, and L.~Zhang, ``Style normalization and restitution for generalizable person re-identification,'' in \emph{CVPR}, 2020.

\bibitem{dai2021generalizable}
Y.~Dai, X.~Li, J.~Liu, Z.~Tong, and L.-Y. Duan, ``Generalizable person re-identification with relevance-aware mixture of experts,'' in \emph{CVPR}, 2021.

\bibitem{zhang2022adaptive}
P.~Zhang, H.~Dou, Y.~Yu, and X.~Li, ``Adaptive cross-domain learning for generalizable person re-identification,'' in \emph{ECCV}, 2022.

\bibitem{jiao2022dynamically}
B.~Jiao, L.~Liu, L.~Gao, G.~Lin, L.~Yang, S.~Zhang, P.~Wang, and Y.~Zhang, ``Dynamically transformed instance normalization network for generalizable person re-identification,'' in \emph{ECCV}, 2022.

\bibitem{xu2022mimic}
B.~Xu, J.~Liang, L.~He, and Z.~Sun, ``Mimic embedding via adaptive aggregation: learning generalizable person re-identification,'' in \emph{ECCV}, 2022.

\bibitem{lu2022augmented}
Y.~Lu, M.~Wang, and W.~Deng, ``Augmented geometric distillation for data-free incremental person reid,'' in \emph{CVPR}, 2022.

\bibitem{ge2022lifelong}
W.~Ge, J.~Du, A.~Wu, Y.~Xian, K.~Yan, F.~Huang, and W.-S. Zheng, ``Lifelong person re-identification by pseudo task knowledge preservation,'' in \emph{AAAI}, 2022.

\bibitem{yu2023lifelong}
C.~Yu, Y.~Shi, Z.~Liu, S.~Gao, and J.~Wang, ``Lifelong person re-identification via knowledge refreshing and consolidation,'' in \emph{AAAI}, 2023.

\bibitem{xu2024lstkc}
K.~Xu, X.~Zou, and J.~Zhou, ``Lstkc: Long short-term knowledge consolidation for lifelong person re-identification,'' in \emph{AAAI}, 2024.

\bibitem{huang2022lifelong}
Z.~Huang, Z.~Zhang, C.~Lan, W.~Zeng, P.~Chu, Q.~You, J.~Wang, Z.~Liu, and Z.-j. Zha, ``Lifelong unsupervised domain adaptive person re-identification with coordinated anti-forgetting and adaptation,'' in \emph{CVPR}, 2022.

\bibitem{Wu2018UnsupervisedFL}
Z.~Wu, Y.~Xiong, S.~X. Yu, and D.~Lin, ``Unsupervised feature learning via non-parametric instance discrimination,'' in \emph{CVPR}, 2018.

\bibitem{He_2020_CVPR}
K.~He, H.~Fan, Y.~Wu, S.~Xie, and R.~Girshick, ``Momentum contrast for unsupervised visual representation learning,'' in \emph{CVPR}, 2020.

\bibitem{chen2020simple}
T.~Chen, S.~Kornblith, M.~Norouzi, and G.~Hinton, ``A simple framework for contrastive learning of visual representations,'' in \emph{ICML}, 2020.

\bibitem{Wang2021camawareproxies}
M.~Wang, B.~Lai, J.~Huang, X.~Gong, and X.-S. Hua, ``Camera-aware proxies for unsupervised person re-identification,'' in \emph{AAAI}, 2021.

\bibitem{Chen_2021_CVPR}
H.~Chen, Y.~Wang, B.~Lagadec, A.~Dantcheva, and F.~Bremond, ``Joint generative and contrastive learning for unsupervised person re-identification,'' in \emph{CVPR}, 2021.

\bibitem{chen2022learning}
------, ``Learning invariance from generated variance for unsupervised person re-identification,'' \emph{IEEE TPAMI}, 2023.

\bibitem{rusu2016progressive}
A.~A. Rusu, N.~C. Rabinowitz, G.~Desjardins, H.~Soyer, J.~Kirkpatrick, K.~Kavukcuoglu, R.~Pascanu, and R.~Hadsell, ``Progressive neural networks,'' \emph{arXiv preprint arXiv:1606.04671}, 2016.

\bibitem{yoon2018lifelong}
J.~Yoon, E.~Yang, J.~Lee, and S.~J. Hwang, ``Lifelong learning with dynamically expandable networks,'' in \emph{ICLR}, 2018.

\bibitem{mallya2018packnet}
A.~Mallya and S.~Lazebnik, ``Packnet: Adding multiple tasks to a single network by iterative pruning,'' in \emph{CVPR}, 2018.

\bibitem{Li2018LearningWF}
Z.~Li and D.~Hoiem, ``Learning without forgetting,'' \emph{IEEE TPAMI}, 2018.

\bibitem{douillard2021plop}
A.~Douillard, Y.~Chen, A.~Dapogny, and M.~Cord, ``Plop: Learning without forgetting for continual semantic segmentation,'' in \emph{CVPR}, 2021.

\bibitem{shang2023incrementer}
C.~Shang, H.~Li, F.~Meng, Q.~Wu, H.~Qiu, and L.~Wang, ``Incrementer: Transformer for class-incremental semantic segmentation with knowledge distillation focusing on old class,'' in \emph{CVPR}, 2023.

\bibitem{Rebuffi2017iCaRLIC}
S.-A. Rebuffi, A.~Kolesnikov, G.~Sperl, and C.~H. Lampert, ``icarl: Incremental classifier and representation learning,'' in \emph{CVPR}, 2017.

\bibitem{Castro2018End}
F.~M. Castro, M.~J. Mar{\'i}n-Jim{\'e}nez, N.~Guil, C.~Schmid, and K.~Alahari, ``End-to-end incremental learning,'' in \emph{ECCV}, 2018.

\bibitem{cha2021co2l}
H.~Cha, J.~Lee, and J.~Shin, ``Co2l: Contrastive continual learning,'' in \emph{ICCV}, 2021.

\bibitem{luo2023class}
Z.~Luo, Y.~Liu, B.~Schiele, and Q.~Sun, ``Class-incremental exemplar compression for class-incremental learning,'' in \emph{CVPR}, 2023.

\bibitem{choi2021dual}
Y.~Choi, M.~El-Khamy, and J.~Lee, ``Dual-teacher class-incremental learning with data-free generative replay,'' in \emph{CVPR}, 2021.

\bibitem{van2020brain}
G.~M. Van~de Ven, H.~T. Siegelmann, and A.~S. Tolias, ``Brain-inspired replay for continual learning with artificial neural networks,'' \emph{Nature communications}, vol.~11, no.~1, p. 4069, 2020.

\bibitem{wang2022continual}
Q.~Wang, O.~Fink, L.~Van~Gool, and D.~Dai, ``Continual test-time domain adaptation,'' in \emph{CVPR}, 2022.

\bibitem{wu2022neighborhood}
S.~Wu, L.~Chen, Y.~Lou, Y.~Bai, T.~Bai, M.~Deng, and L.-Y. Duan, ``Neighborhood consensus contrastive learning for backward-compatible representation,'' in \emph{AAAI}, 2022.

\bibitem{pan2023boundary}
T.~Pan, F.~Xu, X.~Yang, S.~He, C.~Jiang, Q.~Guo, F.~Qian, X.~Zhang, Y.~Cheng, L.~Yang \emph{et~al.}, ``Boundary-aware backward-compatible representation via adversarial learning in image retrieval,'' in \emph{CVPR}, 2023.

\bibitem{oh2024lifelong}
M.~Oh and J.-Y. Sim, ``Lifelong person re-identification with backward-compatibility,'' \emph{arXiv preprint arXiv:2403.10022}, 2024.

\bibitem{Ester1996ADA}
M.~Ester, H.-P. Kriegel, J.~Sander, and X.~Xu, ``A density-based algorithm for discovering clusters in large spatial databases with noise,'' in \emph{KDD}, 1996.

\bibitem{zhong2017re}
Z.~Zhong, L.~Zheng, D.~Cao, and S.~Li, ``Re-ranking person re-identification with k-reciprocal encoding,'' in \emph{CVPR}, 2017.

\bibitem{sohn2020fixmatch}
K.~Sohn, D.~Berthelot, C.-L. Li, Z.~Zhang, N.~Carlini, E.~D. Cubuk, A.~Kurakin, H.~Zhang, and C.~Raffel, ``Fixmatch: Simplifying semi-supervised learning with consistency and confidence,'' in \emph{NeurIPS}, 2020.

\bibitem{sun2019dissecting}
X.~Sun and L.~Zheng, ``Dissecting person re-identification from the viewpoint of viewpoint,'' in \emph{CVPR}, 2019.

\bibitem{Zheng2015ScalablePR}
L.~Zheng, L.~Shen, L.~Tian, S.~Wang, J.~Wang, and Q.~Tian, ``Scalable person re-identification: A benchmark,'' \emph{ICCV}, 2015.

\bibitem{Xiao2017JointDA}
T.~Xiao, S.~Li, B.~Wang, L.~Lin, and X.~Wang, ``Joint detection and identification feature learning for person search,'' \emph{CVPR}, 2017.

\bibitem{wei2018person}
L.~Wei, S.~Zhang, W.~Gao, and Q.~Tian, ``Person transfer gan to bridge domain gap for person re-identification,'' in \emph{CVPR}, 2018.

\bibitem{Gray2008ViewpointIP}
D.~Gray and H.~Tao, ``Viewpoint invariant pedestrian recognition with an ensemble of localized features,'' in \emph{ECCV}, 2008.

\bibitem{hirzer11}
M.~Hirzer, C.~Beleznai, P.~M. Roth, and H.~Bischof, ``{Person Re-Identification by Descriptive and Discriminative Classification},'' in \emph{{Proc. Scandinavian Conference on Image Analysis (SCIA)}}, 2011.

\bibitem{Loy2009MulticameraAC}
C.~C. Loy, T.~Xiang, and S.~Gong, ``Multi-camera activity correlation analysis,'' in \emph{CVPR}, 2009.

\bibitem{Zheng2009AssociatingGO}
W.-S. Zheng, S.~Gong, and T.~Xiang, ``Associating groups of people,'' in \emph{BMVC}, 2009.

\bibitem{Li2012HumanRW}
W.~Li, R.~Zhao, and X.~Wang, ``Human reidentification with transferred metric learning,'' in \emph{ACCV}, 2012.

\bibitem{Li2013LocallyAF}
W.~Li and X.~Wang, ``Locally aligned feature transforms across views,'' \emph{CVPR}, 2013.

\bibitem{Zhao2017SpindleNP}
H.~Zhao, M.~Tian, S.~Sun, J.~Shao, J.~Yan, S.~Yi, X.~Wang, and X.~Tang, ``Spindle net: Person re-identification with human body region guided feature decomposition and fusion,'' \emph{CVPR}, 2017.

\bibitem{Li2014DeepReIDDF}
W.~Li, R.~Zhao, T.~Xiao, and X.~Wang, ``Deepreid: Deep filter pairing neural network for person re-identification,'' \emph{CVPR}, 2014.

\bibitem{3dpes}
D.~Baltieri, R.~Vezzani, and R.~Cucchiara, ``3dpes: 3d people dataset for surveillance and forensics,'' in \emph{Proceedings of the 2011 joint ACM workshop on Human gesture and behavior understanding}, 2011.

\bibitem{PyTorch_NEURIPS2019}
A.~Paszke, S.~Gross, F.~Massa, A.~Lerer, J.~Bradbury, G.~Chanan, T.~Killeen, Z.~Lin, N.~Gimelshein, L.~Antiga, A.~Desmaison, A.~Kopf, E.~Yang, Z.~DeVito, M.~Raison, A.~Tejani, S.~Chilamkurthy, B.~Steiner, L.~Fang, J.~Bai, and S.~Chintala, ``Pytorch: An imperative style, high-performance deep learning library,'' in \emph{NeurIPS}, 2019.

\bibitem{Russakovsky2015ImageNetLS}
O.~Russakovsky, J.~Deng, H.~Su, J.~Krause, S.~Satheesh, S.~Ma, Z.~Huang, A.~Karpathy, A.~Khosla, M.~Bernstein, A.~Berg, and L.~Fei-Fei, ``Imagenet large scale visual recognition challenge,'' \emph{IJCV}, 2015.

\bibitem{he2016deep}
K.~He, X.~Zhang, S.~Ren, and J.~Sun, ``Deep residual learning for image recognition,'' in \emph{CVPR}, 2016.

\bibitem{Zhong2020RandomED}
Z.~Zhong, L.~Zheng, G.~Kang, S.~Li, and Y.~Yang, ``Random erasing data augmentation,'' in \emph{AAAI}, 2020.

\bibitem{Adam_optimization}
D.~P. Kingma and J.~Ba, ``Adam: A method for stochastic optimization,'' in \emph{ICLR}, 2015.

\bibitem{dai2022cluster}
Z.~Dai, G.~Wang, W.~Yuan, S.~Zhu, and P.~Tan, ``Cluster contrast for unsupervised person re-identification,'' in \emph{ACCV}, 2022.

\bibitem{hinton2015distilling}
G.~Hinton, O.~Vinyals, and J.~Dean, ``Distilling the knowledge in a neural network,'' \emph{arXiv preprint arXiv:1503.02531}, 2015.

\bibitem{Han_2023_WACV}
X.~Han, Q.~You, C.~Wang, Z.~Zhang, P.~Chu, H.~Hu, J.~Wang, and Z.~Liu, ``Mmptrack: Large-scale densely annotated multi-camera multiple people tracking benchmark,'' in \emph{WACV}, 2023.

\end{thebibliography}


%

\end{document}